\documentclass[journal]{IEEEtran}

\ifdefined\XeTeXversion
  \usepackage{fontspec}
\fi

\usepackage{amsmath,amssymb,amsfonts,amsthm}
\usepackage{algorithmic}
\usepackage{algorithm}
\usepackage{graphicx}
\usepackage{booktabs}
\usepackage{multirow}
\usepackage{array}
\usepackage{xcolor}
\usepackage{cite}
\usepackage{microtype}
\usepackage[hidelinks]{hyperref}

\newtheorem{proposition}{Proposition}
\newtheorem{remark}{Remark}

\newsavebox{\datatablesave}
\newenvironment{fitdatatable}
  {\begin{lrbox}{\datatablesave}}
  {\end{lrbox}\ifdim\wd\datatablesave>\linewidth
     \resizebox{\linewidth}{!}{\usebox{\datatablesave}}%
   \else\usebox{\datatablesave}\fi}

\newcommand{\R}{\mathbb{R}}
\newcommand{\Sp}{\mathbb{S}}
\newcommand{\norm}[1]{\left\lVert #1 \right\rVert}

\begin{document}

\title{Unlocking Dense Vision Foundation Models for Training-Free Few-Shot Anomaly Detection via Position-Debiased Piecewise-Linear Manifolds}

\author{Jianming~Liu, and
Fan~Xu,~\IEEEmembership{Member,~IEEE}%
\thanks{(Corresponding author: Fan Xu.)}%
\thanks{Jianming Liu is with the School of Artificial
Intelligence, Jiangxi Normal University, Nanchang
330022, China (e-mail: \texttt{liujianming@jxnu.edu.cn}).}%
\thanks{Fan Xu is with the School of Artificial
Intelligence, Jiangxi Normal University, Nanchang
330022, China (e-mail: \texttt{xufan@jxnu.edu.cn}).}%
}
\markboth{Manuscript Submitted to IEEE Transactions on Image Processing}%
{Liu: Position-Debiased Piecewise-Linear Manifolds for Training-Free Few-Shot Anomaly Detection}

\maketitle

\begin{abstract}
Training-free few-shot anomaly detection (FSAD) aims to identify industrial defects using a small number of normal reference images and frozen vision foundation models (VFMs). Although local subspace modeling provides a natural way to characterize heterogeneous normal patterns, our experiments show that directly applying it to DINOv3 features can substantially degrade detection performance. We attribute this degradation primarily to a low-dimensional positional component that dominates feature variance, causing patches to be clustered by spatial location rather than part appearance. This makes local subspace modeling particularly sensitive to object pose variations. To address this problem, we propose Debiased Piecewise-Linear Manifold Anomaly Detection (DPLM-AD), a training-free framework that separates cross-image appearance modeling from intra-image spatial refinement. Specifically, we estimate the positional subspace using images without object-level semantic content and project patch features onto its orthogonal complement. Local affine subspaces are then constructed in the resulting feature space, retaining feature-norm information that may be lost through unit normalization. Instead of discarding positional information entirely, we use the original features to guide graph-based diffusion of anomaly scores, exploiting their spatial continuity for defect localization. Controlled experiments show that positional debiasing changes the effect of local subspace modeling on image-level AUROC from a decrease of 2.87 percentage points to an increase of 0.77 percentage points. Evaluations across 41 categories from five industrial benchmarks demonstrate the effectiveness of DPLM-AD in anomaly detection and localization. On the 27-category MVTec-AD and VisA benchmark suite, DPLM-AD with DINOv3-L outperforms SubspaceAD with DINOv2-G in composite performance while achieving 3.93 times the inference speed.
\end{abstract}

\begin{IEEEkeywords}
Few-shot anomaly detection, vision foundation models, positional debiasing, local subspace modeling, graph diffusion.
\end{IEEEkeywords}

\section{Introduction}
\label{sec:intro}

\IEEEPARstart{A}{utomated} visual anomaly detection (AD) plays an important role in industrial quality inspection, where the objective is to identify defects such as scratches, solder bridges, broken components, and structural abnormalities~\cite{bergmann2019mvtec,zou2022spot,jezek2021mpdd}. Since defective samples are often scarce and difficult to collect, most industrial anomaly detection methods model the distribution of normal samples and identify deviations from this distribution during inference. While conventional approaches rely on a relatively large number of normal training images~\cite{defard2021padim,roth2022patchcore,zavrtanik2021draem,deng2022rd4ad,liu2023simplenet,liang2023omnifreq,li2024target}, collecting such data may be impractical when production lines frequently introduce new products or configurations. Few-shot anomaly detection (FSAD)~\cite{huang2022regad,jeong2023winclip} addresses this problem by using only a small number of normal reference images, typically $k \in \{1,2,4\}$.

Recent advances in vision-language models (VLMs) and self-supervised vision foundation models (VFMs)~\cite{radford2021clip,caron2021dino,oquab2024dinov2} have provided new opportunities for FSAD. Prompt learning and feature adaptation methods~\cite{zhou2024anomalyclip,li2024promptad,zhu2024inctrl,ma2025rempad,jiang2026dcpsfr} exploit pretrained representations but generally require additional optimization or auxiliary training data. In contrast, \emph{training-free} methods construct normality models directly from frozen features. AnomalyDINO~\cite{damm2025anomalydino} and FastRef~\cite{li2026fastref} detect anomalies through reference feature matching or prototype refinement, whereas SubspaceAD~\cite{lendering2026subspacead} models normal patch features using a global principal component analysis (PCA) subspace. Subspace modeling is particularly attractive because it provides a compact representation of normal features without maintaining a large reference memory bank.

\begin{figure*}[t]
    \centering
    \includegraphics[width=\textwidth]{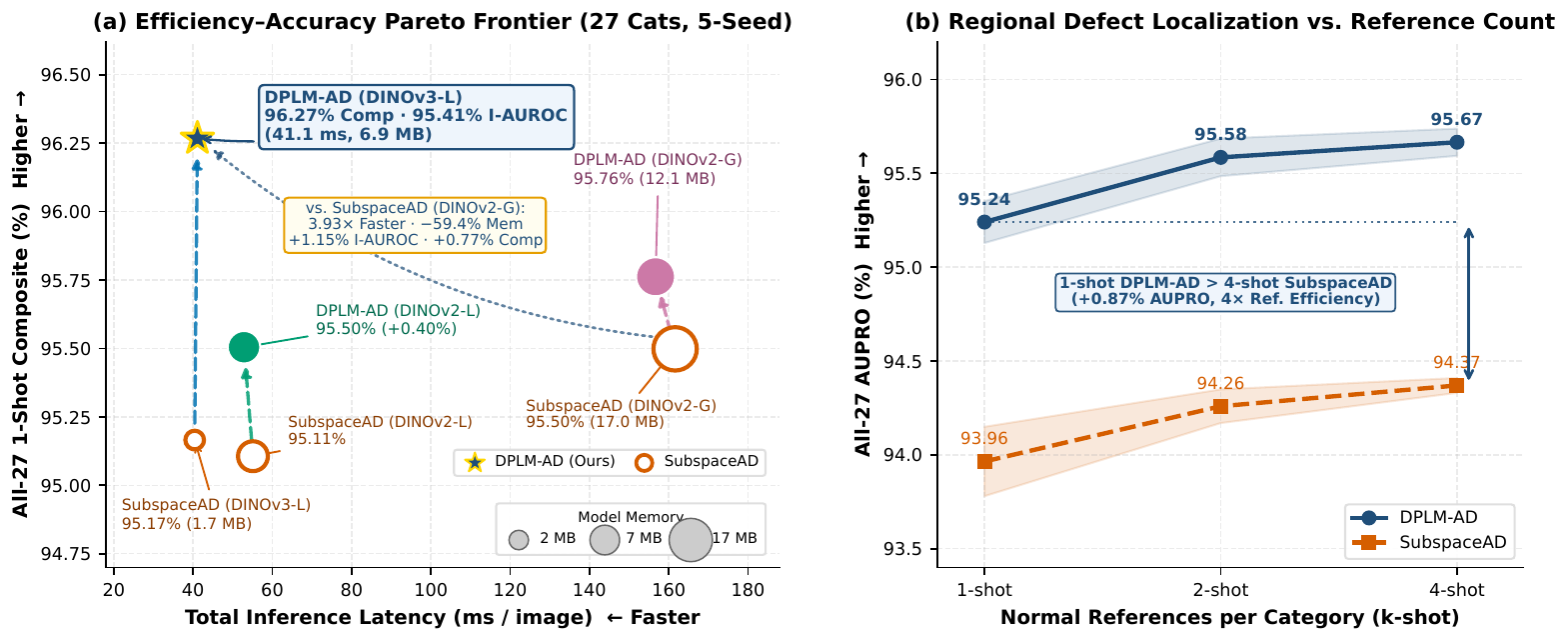}
    \caption{Accuracy and computational efficiency of DPLM-AD on
    the 27 categories of MVTec-AD and VisA.
    (a) Efficiency--accuracy comparison under the 1-shot setting
    over five random seeds. Bubble area represents per-category
    FP32 model storage. On the same DINOv3-L/16 backbone,
    DPLM-AD introduces an additional inference latency of
    $0.69$\,ms. Compared with SubspaceAD using DINOv2-G/14,
    DPLM-AD achieves higher composite performance with
    $3.93\times$ the inference speed.
    (b) Regional anomaly localization performance (AUPRO) under
    different numbers of normal references. With a single reference
    image, DPLM-AD exceeds 4-shot SubspaceAD by $0.87$
    percentage points in AUPRO.}
    \label{fig:teaser}
\end{figure*}

However, a single global linear subspace may not adequately characterize the feature distribution of objects composed of multiple visually distinct parts. Features extracted from different components, such as those in printed circuit boards and mechanical assemblies, may exhibit a multimodal distribution. A global subspace fitted to these features can include regions that are poorly supported by normal observations, potentially reducing its ability to distinguish anomalous patterns. A natural alternative is to approximate the normal feature distribution using multiple local subspaces~\cite{tipping1999mixtures,vidal2011subspace}.

Our experiments, however, reveal an unexpected limitation of this approach. Replacing DINOv2~\cite{oquab2024dinov2} with DINOv3~\cite{simeoni2025dinov3} does not necessarily improve the performance of global subspace modeling, as shown in Fig.~\ref{fig:teaser}(a). More importantly, replacing a single global subspace with $M=12$ local subspaces on raw DINOv3 features reduces image-level AUROC by $2.87$ percentage points across 27 categories from MVTec-AD and VisA. These observations suggest that the effectiveness of local subspace modeling depends not only on the representation capability of the backbone but also on the geometric properties of its feature space.

To understand this behavior, we investigate the positional component of dense DINOv3 features, motivated by recent observations in training-free in-context segmentation~\cite{cuttano2026insid3}. We find that a positional subspace of only 32 dimensions accounts for $81.2\%$ of the centered feature variance in normal test images, despite the original feature dimension being 1024. Clustering in the original feature space therefore tends to group patches according to their spatial locations rather than their visual appearances. For objects with spatially heterogeneous part distributions, changes in object pose may produce large reconstruction residuals even for normal patches. In contrast, this effect is less pronounced for spatially stationary textures. Under matched estimation settings, removing the identified positional component changes the effect of local subspace modeling from a decrease of $2.87$ percentage points to an improvement of $0.77$ percentage points in image-level AUROC. This result indicates that positional bias is an important factor limiting local subspace modeling with dense VFM features.

Nevertheless, positional information is not necessarily undesirable in all stages of anomaly detection. Cross-image normality modeling requires features that are relatively insensitive to changes in object location and pose, whereas anomaly localization within an individual image can benefit from spatial continuity and local feature similarity. Removing positional information from both stages may therefore discard information useful for localization. Another consideration is the geometry of the debiased features. Although unit normalization facilitates cosine-based clustering, it removes feature-norm information that can contribute to distinguishing normal and anomalous patches. These observations motivate the use of different feature representations for normality modeling and spatial refinement.

\begin{figure*}[t]
    \centering
    \includegraphics[width=\textwidth]{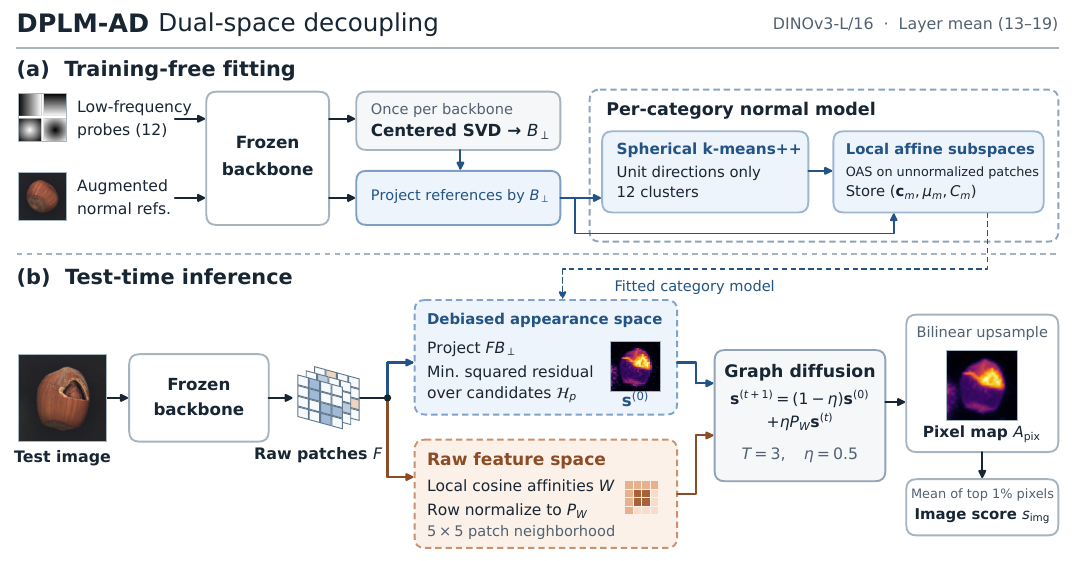}
    \caption{Overview of the DPLM-AD dual-space framework.
    (a) Semantic-free probes define the full-rank positional
    complement $B_\perp$ offline for each backbone.
    For each category, we partition normalized projected
    reference features into $M=12$ semantic modes and fit
    OAS-regularized local Euclidean affine subspaces using
    unnormalized projected features.
    (b) During inference, the debiased branch computes initial
    anomaly scores $\mathbf{s}^{(0)}$ from the minimum squared
    affine residual~\eqref{eq:uoa_score}. The raw-feature branch
    constructs boundary-aware transition weights $P_W$ for
    row-stochastic graph diffusion, followed by bilinear
    upsampling and top-$1\%$ pixel TVaR pooling.}
    \label{fig:dual_space_decoupling}
\end{figure*}

Based on these findings, we propose \textbf{DPLM-AD} (\textbf{D}ebiased \textbf{P}iecewise-\textbf{L}inear \textbf{M}anifold \textbf{A}nomaly \textbf{D}etection), a training-free framework that separates cross-image appearance modeling from intra-image anomaly score refinement. As illustrated in Fig.~\ref{fig:dual_space_decoupling}, we first estimate a positional subspace from probe images without object-level semantic content and project the reference features onto its orthogonal complement $B_\perp \in \R^{D \times (D-s)}$. The projected features are partitioned using spherical $k$-means++~\cite{banerjee2005clustering}, while local affine subspaces are estimated from the unnormalized projected features with Oracle Approximating Shrinkage (OAS)~\cite{chen2010shrinkage}. Anomaly scores are computed from the minimum squared reconstruction residual over candidate local subspaces. Instead of using the projected features throughout the entire pipeline, we retain the original features to construct a local affinity graph and refine anomaly scores through row-stochastic graph diffusion~\cite{grady2006random,ortega2018graph}. This design reduces the influence of positional bias on cross-image normality modeling while preserving information useful for intra-image localization.

We evaluate DPLM-AD across 41 categories from five industrial anomaly detection benchmarks: MVTec-AD, VisA, MPDD, BTAD, and MVTec-LOCO. The experimental results show improvements in anomaly detection and localization across the structural anomaly benchmarks with limited additional computational overhead. As shown in Fig.~\ref{fig:teaser}, DPLM-AD with DINOv3-L/16 achieves higher composite performance than SubspaceAD with DINOv2-G/14 on MVTec-AD and VisA while providing $3.93\times$ the inference speed. We also examine the limitations of local subspace modeling when normal samples are insufficient and when anomalies depend on global spatial arrangements.

The main contributions of this work are threefold:
\begin{itemize}
    \item \textbf{Analysis of positional bias in local subspace modeling.}
    We investigate how a low-dimensional positional component in dense
    VFM features affects feature clustering and subspace-based anomaly
    detection. Through theoretical analysis, real-feature diagnostics,
    and controlled experiments under different pose conditions, we
    explain why local subspace modeling can degrade detection
    performance and identify conditions under which positional
    information is detrimental or beneficial.

    \item \textbf{Position-debiased local manifold modeling.}
    We develop a training-free normality model that combines orthogonal
    positional projection, cosine-based partitioning, and
    shrinkage-regularized local affine subspaces. The model retains
    feature-norm information during subspace estimation and anomaly
    scoring to characterize heterogeneous normal feature distributions.
    We further investigate how the effectiveness of local modeling
    depends on feature multimodality and sample availability.

    \item \textbf{Dual-space anomaly detection and localization.}
    We introduce a framework that uses position-debiased features for
    cross-image normality modeling and raw features for intra-image
    graph diffusion. Controlled comparisons demonstrate the complementary
    roles of these representations. Experiments across five industrial
    anomaly detection benchmarks demonstrate the effectiveness and
    computational efficiency of the proposed framework, while also
    revealing its limitations on certain logical anomalies.
\end{itemize}

\section{Related Work}
\label{sec:related}

\subsection{Unsupervised and Few-Shot Industrial Anomaly Detection}
Full-shot AD methods model normal distributions via Gaussian estimators~\cite{defard2021padim}, coreset memory banks~\cite{roth2022patchcore,li2024target}, synthetic anomaly reconstruction~\cite{zavrtanik2021draem,liu2023simplenet}, or reverse distillation~\cite{deng2022rd4ad,liang2023omnifreq}. Under strict few-shot constraints ($k \le 4$), early work used Siamese registration~\cite{huang2022regad}, while VLM pipelines introduced compositional prompting (WinCLIP~\cite{jeong2023winclip}), prompt tuning (AnomalyCLIP~\cite{zhou2024anomalyclip}, PromptAD~\cite{li2024promptad}), in-context residual learning (InCTRL~\cite{zhu2024inctrl}), and auxiliary-trained adaptation (ReMP-AD~\cite{ma2025rempad}, DCP-SFR~\cite{jiang2026dcpsfr}). Purely \emph{training-free} VFM frameworks avoid external defect data and backpropagation: AnomalyDINO~\cite{damm2025anomalydino} and FastRef~\cite{li2026fastref} perform patch retrieval or optimal-transport prototype refinement, whereas SubspaceAD~\cite{lendering2026subspacead} fits a global PCA subspace with Gaussian smoothing and Tail Value-at-Risk (TVaR)~\cite{rockafellar2000optimization} pooling. DPLM-AD retains this closed-form, memory-bank-free paradigm while overcoming single-plane over-generalization and DINOv3 positional collapse.

\subsection{Positional Bias, Local Manifolds, and Graph Diffusion}
Dense self-supervised Vision Transformers~\cite{caron2021dino,oquab2024dinov2,darcet2024registers} provide rich part correspondence; DINOv3~\cite{simeoni2025dinov3} pairs rotary encodings~\cite{su2024roformer} with Gram anchoring. In training-free segmentation, INSID3~\cite{cuttano2026insid3} identified a low-dimensional positional subspace from unstructured probe images and projected it out in ambient $\R^D$ for cosine matching. We advance beyond ambient retrieval by: (i) formulating full-rank isometric compression into $\R^{D-s}$ prior to OAS-regularized local Euclidean covariance estimation~\cite{tipping1999mixtures,vidal2011subspace,banerjee2005clustering,ledoit2004well,chen2010shrinkage}; (ii) quantifying how positional bias hijacks local clustering on real features and across pose regimes; and (iii) repurposing raw features to drive edge-preserving random-walk graph diffusion~\cite{perona1990scalespace,he2013guided,zhou2003learning,grady2006random,shuman2013emerging,ortega2018graph} rather than isotropic Gaussian blurring.

\section{Methodology}
\label{sec:method}

\subsection{Problem Formulation and Multi-Layer Aggregation}

Let $\Phi(\cdot)$ denote a frozen Vision Transformer backbone
(\texttt{DINOv3-L/16} by default). Given $k \in \{1, 2, 4\}$
normal references
$\mathcal{I}_{\text{ref}} = \{I_1, \dots, I_k\}$
and a test image $I_{\text{test}} \in \R^{H \times W \times 3}$,
training-free FSAD produces a pixel-level anomaly map
$A_{\text{pix}} \in \R^{H \times W}$ and an image-level
anomaly score $s_{\text{img}} \in \R$.

After removing the \texttt{[CLS]} and register
tokens~\cite{darcet2024registers}, we average the patch
tokens from a set of intermediate layers $\mathcal{L}$:
\begin{equation}
\mathbf{x}_p = \frac{1}{|\mathcal{L}|}
\sum_{l \in \mathcal{L}} \mathbf{f}_l(p) \in \R^{D},
\quad p \in \{1, \dots, P\},
\label{eq:layer_agg}
\end{equation}
where $P = h \times w$ and
$F = [\mathbf{x}_1, \dots, \mathbf{x}_P]^\top
\in \R^{P \times D}$.
Applying $N_a$ rotation augmentations to orientation-invariant
classes yields $N = k(1+N_a)P$ normal patch vectors
$\{\mathbf{x}_i\}_{i=1}^N$.

\subsection{Offline Full-Rank Orthogonal Positional Debiasing}

We estimate the dominant positional subspace once for each
backbone and input resolution using $N_{\text{pr}} = 12$
semantic-free low-frequency probes
$\{I_j^{\text{probe}}\}_{j=1}^{12}$.
These probes comprise nine analytic fields (three uniform
grayscale images, four linear gradients, and two radial
gradients) and three heavily smoothed, low-amplitude fields
generated with fixed random seeds.

We extract their features using $\Phi$ and the layer set
$\mathcal{L}$, yielding
$F^{\text{probe}} \in \R^{(N_{\text{pr}}P) \times D}$.
We then center $F^{\text{probe}}$ along the sample dimension
and apply singular value decomposition (SVD):
\begin{equation}
\bar{F}^{\text{probe}}
= F^{\text{probe}} - \mathbf{1}\boldsymbol{\mu}_{\text{probe}}^\top
= U \Sigma V^\top,
\quad V = [\mathbf{v}_1, \dots, \mathbf{v}_D].
\label{eq:probe_svd}
\end{equation}

Let $B = [\mathbf{v}_1, \dots, \mathbf{v}_s]
\in \R^{D \times s}$ span the top-$s$ positional directions,
and let $B_\perp = [\mathbf{v}_{s+1}, \dots, \mathbf{v}_D]
\in \R^{D \times d}$ ($d = D - s$) form an orthonormal basis
for its complement.
Instead of retaining the ambient-dimensional representation
obtained with the rank-deficient projector $I_D - BB^\top$,
we express the projected features directly in the
$d$-dimensional complementary space:
\begin{equation}
\hat{\mathbf{x}}_p = B_\perp^\top \mathbf{x}_p \in \R^{d},
\qquad
\hat{F} = F B_\perp \in \R^{P \times d}.
\label{eq:full_rank_proj}
\end{equation}

As shown in Section~\ref{sec:exp_ablation}, the estimated
32-dimensional subspace $B$ is consistent across
semantic-free probe families of equal size, with a top-8
Grassmann cosine of $0.971$ relative to the subspace
estimated from 12 white-noise probes.
On unseen normal test images, $B$ accounts for $81.2\%$
of the centered feature variance and $88.8\%$ of the
spatial-grid variance.

\begin{proposition}[Isometry and Rank Preservation]
\label{prop:isometry}
For any $\mathbf{x}_p \in \R^D$ with ambient projection
$\tilde{\mathbf{x}}_p = (I_D - BB^\top)\mathbf{x}_p$,
the complementary representation
$\hat{\mathbf{x}}_p = B_\perp^\top \mathbf{x}_p \in \R^d$
satisfies:
(i) strict isometry,
$\langle \hat{\mathbf{x}}_p, \hat{\mathbf{x}}_q \rangle
= \langle \tilde{\mathbf{x}}_p, \tilde{\mathbf{x}}_q \rangle$
and
$\norm{\hat{\mathbf{x}}_p}_2
= \norm{\tilde{\mathbf{x}}_p}_2$;
and (ii) preservation of covariance rank while eliminating
the $s$ redundant dimensions associated with $\mathrm{col}(B)$,
reducing the eigendecomposition cost from $\mathcal{O}(D^3)$
to $\mathcal{O}((D-s)^3)$.
\end{proposition}

\begin{proof}
Since $V=[B, B_\perp]$ is orthonormal,
$I_D - BB^\top = B_\perp B_\perp^\top$ and
$B_\perp^\top B_\perp = I_d$.
It follows that
$\tilde{\mathbf{x}}_p^\top \tilde{\mathbf{x}}_q
= \hat{\mathbf{x}}_p^\top \hat{\mathbf{x}}_q$.
The ambient covariance
$\tilde{\Sigma} = B_\perp \hat{\Sigma} B_\perp^\top$
vanishes on $\mathrm{col}(B)$.
Since $B_\perp$ has orthonormal columns,
$\mathrm{rank}(\tilde{\Sigma})
= \mathrm{rank}(\hat{\Sigma})$.
The complementary representation thus removes the $s$
structural zero eigenvalues introduced by the ambient
projection.
\end{proof}

Here, full rank refers to the projection basis $B_\perp$,
rather than the empirical covariance matrix, which may
remain rank-deficient when the number of samples is limited.

DPLM-AD retains two feature representations for different
purposes: $\hat{F}$ for cross-image local manifold modeling
(Section~\ref{sec:manifold}) and raw $F$ for intra-image
anomaly score diffusion (Section~\ref{sec:diffusion}).

\subsection{Position-Debiased Piecewise-Linear Manifold Modeling}
\label{sec:manifold}

\subsubsection{Positional Bias in Local Clustering:
Theory and Empirical Verification}

To analyze the effect of positional bias on local clustering,
we decompose each raw patch feature as
$\mathbf{x}_i = \mathbf{x}_i^{\text{sem}}
+ \mathbf{x}_i^{\text{pos}}$, where
$\mathbf{x}_i^{\text{sem}} = B_\perp B_\perp^\top \mathbf{x}_i$
and $\mathbf{x}_i^{\text{pos}} = BB^\top \mathbf{x}_i$.
The corresponding unit vector is
$\bar{\mathbf{x}}_i = \mathbf{x}_i / \norm{\mathbf{x}_i}_2
= \bar{\mathbf{x}}_i^{\text{sem}}
+ \bar{\mathbf{x}}_i^{\text{pos}}$.
The superscripts ``sem'' and ``pos'' denote the components
in the complementary and estimated positional subspaces,
respectively, without implying complete disentanglement
of semantic and positional information.

\begin{proposition}[Fixed-Partition Objective Decomposition
and Idealized Constant-Quadrant Limit]
\label{prop:pos_corruption}

For any fixed partition
$\mathcal{P} = \{\mathcal{C}_1, \dots, \mathcal{C}_M\}$
of $\{\mathbf{x}_i\}_{i=1}^N \subset \R^D$ into $M$ clusters:

\begin{enumerate}
    \item \textbf{Orthogonal Objective Decompositions:}
    The Euclidean $k$-means distortion decomposes as
    $J_{\text{euc}}(\mathcal{P})
    = J_{\text{sem}}(\mathcal{P}) + J_{\text{pos}}(\mathcal{P})$.
    A spatial partition $\mathcal{P}_{\text{pos}}$ therefore
    has lower distortion than a semantic partition
    $\mathcal{P}_{\text{sem}}$ if and only if
    $J_{\text{pos}}(\mathcal{P}_{\text{sem}})
    - J_{\text{pos}}(\mathcal{P}_{\text{pos}})
    > J_{\text{sem}}(\mathcal{P}_{\text{pos}})
    - J_{\text{sem}}(\mathcal{P}_{\text{sem}})$.

    For spherical $k$-means
    ($\min \sum_m \sum_{i\in\mathcal{C}_m}
    (1 - \langle \bar{\mathbf{x}}_i, \mathbf{c}_m \rangle)$),
    optimizing the unit centroids
    $\mathbf{c}_m \in \Sp^{D-1}$ yields the equivalent
    maximization objective
    \begin{equation}
    \mathcal{O}_{\text{sph}}(\mathcal{P})
    = \sum_{m=1}^M
    \sqrt{
    \norm{\sum_{i \in \mathcal{C}_m}
    \bar{\mathbf{x}}_i^{\text{sem}}}_2^2
    +
    \norm{\sum_{i \in \mathcal{C}_m}
    \bar{\mathbf{x}}_i^{\text{pos}}}_2^2}.
    \label{eq:sph_obj}
    \end{equation}

    \item \textbf{Idealized Constant-Quadrant Limit:}
    Assume that
    $\mathbf{x}_i^{\text{pos}} \equiv \mathbf{p}_m
    \in \mathrm{col}(B)$ is constant within each
    cluster $\mathcal{C}_m$.
    Let $(\boldsymbol{\mu}_m, C_m)$ denote the
    OAS-shrunk PCA subspace in $\R^D$, with
    $\gamma_m \in (0, 1)$ and
    $r_m \le \mathrm{rank}(\hat{\Sigma}_m)$.
    Then $B^\top C_m = \mathbf{0}$.

    For any test patch
    $\mathbf{x}_{\text{test}}
    = \mathbf{x}^{\text{sem}} + \mathbf{p}(\mathbf{u}')$,
    define
    $e_m = \norm{(I_D - C_m C_m^\top)
    (\mathbf{x}^{\text{sem}}
    - \boldsymbol{\mu}_m^{\text{sem}})}_2^2$.
    Over a nonempty candidate set $\mathcal{J}$, the
    minimum squared residual is
    \begin{equation}
    s_{\mathcal{J}}(\mathbf{x}_{\text{test}})
    = \min_{m\in\mathcal{J}}
    \left(e_m +
    \norm{\mathbf{p}(\mathbf{u}')-\mathbf{p}_m}_2^2\right).
    \label{eq:double_bind}
    \end{equation}
\end{enumerate}
\end{proposition}

Supplementary Section~S2 provides the algebraic proofs
and the corresponding triangle-inequality sufficient
condition.
Part~(1) compares fixed partitions and does not guarantee
convergence to the global $k$-means optimum, whereas
Part~(2) assumes exact positional constancy within each
cluster.
To assess how these results apply to real features, we
analyze cached \texttt{DINOv3-L/16} training features
from all 27 MVTec-AD and VisA categories
(1-shot, \texttt{seed=0}; Supplementary Table~S1).

\begin{itemize}
    \item \emph{Positional dominance in clustering:}
    The subspace $\mathrm{col}(B)$ accounts for
    $79.4 \pm 2.2\%$ of the total centered feature variance.
    Centering the probe features does not make the means
    of real features orthogonal to $B$:
    $36.4 \pm 6.9\%$ of
    $\norm{\boldsymbol{\mu}_{\text{real}}}_2^2$
    lies in $\mathrm{col}(B)$.
    Clustering raw features with $M=12$ removes
    $70.5\%$ of the positional variance through cluster
    means alone.
    The within-cluster positional residual ratio is
    $29.5 \pm 2.9\%$, compared with $32.8\%$ for a
    $3\times 4$ spatial grid and $79.2\%$ for the
    debiased partition with $M=12$.
    The raw-feature partition achieves better objective
    values than the debiased partition in both
    $J_{\text{euc}}$ and $\mathcal{O}_{\text{sph}}$
    ($0.861$ vs.\ $0.723$) across all \textbf{$27/27$}
    categories.

    \item \emph{Deviation from the idealized residual
    model in~\eqref{eq:double_bind}:}
    The remaining $29.5\%$ of positional variance accounts
    for $63.8\%$ of the total within-cluster variance.
    Consequently, the local subspaces fitted to raw
    features do not satisfy $B^\top C_m = \mathbf{0}$
    exactly.
    They capture $72.4\%$ of the top-8 positional directions,
    compared with $80.6\%$ for global $M=1$ PCA, and
    $52.0\%$ of the 32-dimensional subspace $B$.
    The remaining $27.6\%$ of the top-8 directions
    ($48.0\%$ of the full 32-dimensional subspace)
    contribute to residuals outside $C_m$, while the
    retained positional components reduce the capacity
    available for modeling semantic variation.
\end{itemize}

\begin{remark}[Three Operational Pose Regimes]
\label{rem:three_regimes}

Equation~\eqref{eq:double_bind} and the empirical
observations above suggest three pose-dependent regimes,
which we evaluate in Section~\ref{sec:exp_counterfactual}:
(i) for \emph{non-stationary multi-part objects} undergoing
pose shifts, matching part appearance incurs a positional
penalty, whereas matching spatial location incurs a
semantic mismatch $e_m$;
(ii) for \emph{spatially stationary textures}, spatial
regions share similar appearance subspaces, allowing
anomaly rankings to remain stable despite changes in
cluster assignments; and
(iii) for \emph{strictly aligned objects} at zero shift
($\mathbf{u}' = \mathbf{u}$),
the positional term
$\norm{\mathbf{p}(\mathbf{u}')-\mathbf{p}_m}_2^2$
provides a useful spatial prior.
As misalignment increases, the position-debiased
representation based on $B_\perp$ can become more
effective.
\end{remark}

\subsubsection{Cosine Partitioning, Local OAS Subspaces,
and Scoring}

We partition the debiased reference features
$\{\hat{\mathbf{x}}_i\}_{i=1}^N$ into $M$ clusters
$\{\mathcal{C}_1, \dots, \mathcal{C}_M\}$ using
spherical $k$-means++~\cite{banerjee2005clustering}
on unit vectors
$\hat{\mathbf{x}}_i / \norm{\hat{\mathbf{x}}_i}_2$.
We enforce a minimum cluster size of $n_{\min}=5$
by splitting the largest clusters and merging undersized
clusters.
We use unit normalization only for clustering and fit
the local subspaces to unnormalized features in $\R^d$
to preserve feature-norm information.

For cluster $m$ ($n_m = |\mathcal{C}_m|$), let
$\boldsymbol{\mu}_m
= \frac{1}{n_m}\sum_{i \in \mathcal{C}_m}
\hat{\mathbf{x}}_i$ denote its mean and
$\hat{\Sigma}_m
= \frac{1}{\max(n_m-1, 1)}
\sum_{i \in \mathcal{C}_m}
(\hat{\mathbf{x}}_i - \boldsymbol{\mu}_m)
(\hat{\mathbf{x}}_i - \boldsymbol{\mu}_m)^\top$
its sample covariance.
To regularize the covariance spectrum for rank selection,
particularly when $n_m < d$, we apply closed-form
Oracle Approximating Shrinkage
(OAS)~\cite{chen2010shrinkage}:
\begin{equation}
\tilde{\Sigma}_m
= (1 - \gamma_m)\hat{\Sigma}_m
+ \gamma_m \frac{\mathrm{Tr}(\hat{\Sigma}_m)}{d} I_d,
\label{eq:oas_cov}
\end{equation}
\begin{equation}
\resizebox{0.88\linewidth}{!}{$\displaystyle
\gamma_m = \mathrm{clip}\!\left(
\frac{
(1 - 2/d)\,\mathrm{Tr}(\hat{\Sigma}_m^2)
+ \mathrm{Tr}(\hat{\Sigma}_m)^2
}{
(n_m + 1 - 2/d)
\left(
\mathrm{Tr}(\hat{\Sigma}_m^2)
- \mathrm{Tr}(\hat{\Sigma}_m)^2 / d
\right)
},\; \gamma_{\min},\; \gamma_{\max}\right).$}
\label{eq:oas_gamma}
\end{equation}

Let $\lambda_{m,1} \ge \dots \ge \lambda_{m,d} > 0$
be the ordered eigenvalues of $\tilde{\Sigma}_m$.
We form the local subspace basis
$C_m \in \R^{d \times r_m}$ from the leading $r_m$
eigenvectors, where
\begin{equation}
r_m = \min\!\left\{
r_{\tau,m},\; r_{\max},\; \max(1, n_m - 2)
\right\},
\label{eq:rank_sel}
\end{equation}
and
$r_{\tau,m}
= \min\bigl\{r :
\sum_{j=1}^r \lambda_{m,j}
\ge \tau \sum_{j=1}^d \lambda_{m,j}\bigr\}$.

Since isotropic OAS shrinkage preserves the eigenspaces
and eigenvalue ordering of $\hat{\Sigma}_m$, its role
in the proposed reconstruction-based model is to regularize
the eigenvalue spectrum used for rank selection rather
than to modify the principal directions.
For a fixed rank with a uniquely defined principal
subspace, the reconstruction residual remains unchanged.

Given the unit centroids $\mathbf{c}_m \in \Sp^{d-1}$
and cluster means $\boldsymbol{\mu}_m \in \R^d$,
we define the stabilized cosine similarity
$a_{pm}
= \langle \hat{\mathbf{x}}_p, \mathbf{c}_m \rangle
/ (\norm{\hat{\mathbf{x}}_p}_2 + 10^{-8})$
for test patch $p$.
Let $m_p^\star = \arg\max_m a_{pm}$
(with ties resolved by the smallest index) and
$\mathbf{v}_{pm}
= \hat{\mathbf{x}}_p - \boldsymbol{\mu}_m$.
We retain clusters with positive cosine similarity
and include $m_p^\star$ as a fallback to ensure a
nonempty candidate set.
The initial anomaly score is the minimum squared
reconstruction residual over this set:
\begin{equation}
\begin{aligned}
\mathcal{H}_p &= \{m : a_{pm}>0\}\cup\{m_p^\star\},\\
s_p^{(0)} &= \min_{m\in\mathcal{H}_p}
    \norm{\mathbf{v}_{pm}
    -C_m(C_m^\top\mathbf{v}_{pm})}_2^2 .
\end{aligned}
\label{eq:uoa_score}
\end{equation}

Across all 405 paired MVTec/VisA runs
($7.16\times 10^8$ patch--cluster pairs),
$a_{pm} > 0$ for every pair, yielding $100\%$ retention
and a $0.0000\%$ metric discrepancy relative to
unrestricted minimization over all $M$ clusters.
Thus, the hemisphere constraint serves only as a
safeguard and does not affect accuracy or runtime
under the evaluated settings.

\subsection{Raw-Feature-Guided Edge-Preserving Graph Diffusion}
\label{sec:diffusion}

Gaussian smoothing ($\sigma=4$) does not account for
local feature similarity.
We instead construct a sparse affinity graph
$W \in \R^{P \times P}$ over $w_g \times w_g$
spatial neighborhoods using normalized \emph{raw}
test features
$\bar{\mathbf{f}}_p
= \mathbf{x}_p^{\text{test}}
/ \norm{\mathbf{x}_p^{\text{test}}}_2$:
\begin{equation}
W_{pq} = \begin{cases}
\exp\!\left(
\dfrac{\langle \bar{\mathbf{f}}_p,
\bar{\mathbf{f}}_q \rangle - 1}{\tau_g}
\right),
& q \in \mathcal{N}_{w_g}(p), \\
0, & \text{otherwise},
\end{cases}
\label{eq:affinity}
\end{equation}
where $\mathcal{N}_{w_g}(p)$ denotes the valid
in-bounds $w_g \times w_g$ neighborhood of patch $p$,
and $\tau_g > 0$ controls the sensitivity of the
affinity weights to feature similarity.

Equation~\eqref{eq:affinity} defines the symmetric
valid-neighbor formulation. For computational efficiency,
our default implementation uses replicate padding,
which preserves row-stochasticity after normalization
but may introduce asymmetry at image boundaries.

We obtain the row-stochastic transition matrix
$P_W = D_W^{-1} W$, where
$D_{W,pp} = \sum_q W_{pq}$.
Starting from $\mathbf{s}^{(0)}$, we perform $T$
damped random-walk iterations with $\eta \in (0, 1)$:
\begin{equation}
\mathbf{s}^{(t+1)}
= (1 - \eta)\,\mathbf{s}^{(0)}
+ \eta\, P_W \mathbf{s}^{(t)},
\quad t = 0, \dots, T-1.
\label{eq:diffusion}
\end{equation}

\begin{proposition}[Contraction, Amplitude Bounds,
and Variational Optimality]
\label{prop:diffusion}

For any $\eta \in (0, 1)$ and nonnegative local
affinity matrix $W$ with positive degrees
($D_{W,pp} > 0$):

\begin{enumerate}
    \item Row-stochasticity
    ($P_W \mathbf{1} = \mathbf{1}$)
    guarantees strict $\ell_\infty$ contraction with
    rate $\eta$ and geometric convergence to
    $\mathbf{s}^*
    = (1-\eta)(I_P - \eta P_W)^{-1}\mathbf{s}^{(0)}$.
    The diffusion preserves constant fields and satisfies
    $\min_q s_q^{(0)} \le s_p^{(t)}
    \le \max_q s_q^{(0)}$ for all $t$.

    \item If $W$ is symmetric ($W_{pq} = W_{qp}$),
    as obtained through valid-neighbor boundary masking,
    $\mathbf{s}^*$ is the unique minimizer of the
    quadratic graph energy
    \begin{equation}
    \begin{aligned}
    \mathcal{J}(\mathbf{s})
    &= \frac{1}{2}\sum_{p,q} W_{pq}(s_p - s_q)^2 \\
    &\quad + \frac{1-\eta}{\eta}
    \sum_p D_{W,pp}\bigl(s_p - s_p^{(0)}\bigr)^2.
    \end{aligned}
    \label{eq:graph_obj}
    \end{equation}

    \item Across a semantic or defect boundary with
    $\langle \bar{\mathbf{f}}_p,
    \bar{\mathbf{f}}_q \rangle \ll 1$,
    $W_{pq} \to 0$ exponentially in $1/\tau_g$.
\end{enumerate}
\end{proposition}

\begin{proof}
Nonnegativity and $P_W\mathbf{1}=\mathbf{1}$ imply
$\norm{\eta P_W}_\infty = \eta < 1$.
The update in~\eqref{eq:diffusion} is therefore a
contraction, and its convex-combination structure
preserves the bounds of the initial scores at every
iteration, including under replicate border padding.

When $W_{pq} = W_{qp}$, the first-order optimality
condition at $\mathbf{s}^*$ is
$\nabla_{\mathbf{s}}\mathcal{J}(\mathbf{s}^*)
= 2\bigl[(D_W - W)\mathbf{s}^*
+ \frac{1-\eta}{\eta}
D_W(\mathbf{s}^* - \mathbf{s}^{(0)})\bigr]
= \mathbf{0}$.
The graph energy is strictly convex because the graph
Laplacian is positive semidefinite and the fidelity term
has positive weights. Hence, $\mathbf{s}^*$ is its
unique minimizer.
The exponential decay of cross-boundary affinity weights
follows directly from~\eqref{eq:affinity}.
\end{proof}

The default replicate-padded implementation preserves
symmetry between interior pixels and satisfies
Property~(1) throughout the feature grid.
Property~(2) is guaranteed only under symmetric affinities.
As shown in Table~\ref{tab:ablation_matched}(a),
valid-neighbor masking (\texttt{A4-sym}) ensures
global symmetry, satisfies Property~(2) over the
entire grid, and slightly improves all metrics
($+0.055\%$ I-AUROC, $+0.024\%$ Comp.).

Finally, we bilinearly upsample $\mathbf{s}^{(T)}$
to obtain $A_{\text{pix}} \in \R^{H \times W}$
and apply top-$1\%$ pixel TVaR pooling
($2{,}621$ pixels at $512^2$) to compute
$s_{\text{img}}$.
Algorithm~\ref{alg:dplm} summarizes the complete
fitting and inference procedure.

\begin{algorithm}[t]
\caption{DPLM-AD: Training-Free Fitting and Inference}
\label{alg:dplm}
\footnotesize
\begin{algorithmic}[1]

\REQUIRE Frozen VFM $\Phi$, layer set $\mathcal{L}$,
probes $\{I_j^{\text{probe}}\}_{j=1}^{12}$,
$k$ normal references $\mathcal{I}_{\text{ref}}$,
test image $I_{\text{test}}$;
defaults $s=32, M=12, \tau=0.95, r_{\max}=192,
w_g=5, \tau_g=0.1, \eta=0.5, T=3$.

\STATE \textbf{Offline Stage (Once per Backbone):}
Extract $F^{\text{probe}}$ using~\eqref{eq:layer_agg}
and apply centered SVD~\eqref{eq:probe_svd}
to obtain $B_\perp \in \R^{D \times (D-s)}$.

\STATE \textbf{Category Fitting Stage (Training-Free):}

\STATE Extract augmented normal patch features
$\{\mathbf{x}_i\}_{i=1}^N$ and project them as
$\hat{\mathbf{x}}_i = B_\perp^\top \mathbf{x}_i$.

\STATE Apply spherical $k$-means++ ($n_{\min}=5$)
to normalized features
$\{\hat{\mathbf{x}}_i/\norm{\hat{\mathbf{x}}_i}_2\}$
to obtain clusters $\{\mathcal{C}_m\}_{m=1}^M$
and unit centroids $\{\mathbf{c}_m\}_{m=1}^M$.

\FOR{$m = 1$ \TO $M$}
    \STATE Compute the mean $\boldsymbol{\mu}_m$
    from unnormalized features, estimate the OAS covariance
    $\tilde{\Sigma}_m$ using
    \eqref{eq:oas_cov}--\eqref{eq:oas_gamma},
    and retain the leading $r_m$ eigenvectors
    $C_m \in \R^{d \times r_m}$
    according to~\eqref{eq:rank_sel}.
\ENDFOR

\STATE \textbf{Online Inference Stage:}

\STATE Extract raw features $F^{\text{test}} \in \R^{P \times D}$,
compute $\hat{F}^{\text{test}} = F^{\text{test}} B_\perp$,
and obtain initial residuals $\mathbf{s}^{(0)}$
using~\eqref{eq:uoa_score}
with candidate set $\mathcal{H}_p$ and fallback $m_p^\star$.

\STATE Construct $W$ from raw $F^{\text{test}}$
using~\eqref{eq:affinity} with the default boundary
handling described in Section~\ref{sec:diffusion},
and compute $P_W = D_W^{-1}W$.
Apply~\eqref{eq:diffusion} for $T$ iterations
to obtain $\mathbf{s}^{(T)}$.

\STATE Bilinearly upsample $\mathbf{s}^{(T)}$
to obtain $A_{\text{pix}}$ and average its top-$1\%$
pixel scores to compute $s_{\text{img}}$.

\RETURN $A_{\text{pix}}, s_{\text{img}}$.

\end{algorithmic}
\end{algorithm}

\subsection{Computational and Storage Complexity}

The offline probe SVD runs once per backbone and
input resolution and produces a shared basis $B_\perp$,
which requires $3.88$\,MB in FP32 or $1.94$\,MB in FP16.
The amortized FP16 storage is $<0.06$\,MB per category
across 36 categories.

Given extracted reference features, the projection costs
$\mathcal{O}(NDd)$.
Under standard dense implementations, the subsequent
per-category fitting has a computational cost of
$\mathcal{O}(T_{\text{km}} M N d + N d^2 + M d^3)$,
where $T_{\text{km}}$ denotes the number of spherical
$k$-means iterations. The three terms correspond to
clustering, covariance estimation, and eigendecomposition,
respectively.
Per-category fitting takes $0.45\text{--}0.85$\,s on
a single GPU.

The inference state consists of
$\{(\mathbf{c}_m, \boldsymbol{\mu}_m, C_m)\}_{m=1}^M$,
requiring $6.85$\,MB in FP32
($6.91$\,MB including auxiliary metadata in
Tables~\ref{tab:ablation_matched}
and~\ref{tab:backbones_5seed})
or \textbf{$3.43$\,MB in FP16}.
For comparison, SubspaceAD with the same backbone
requires $1.67$\,MB in FP32.

The computational cost of post-backbone inference is
$\mathcal{O}(P D d + P M d r_{\max} + T P w_g^2)$.
The measured post-backbone latency is $6.01$\,ms,
with an additional $0.69$\,ms in total inference
latency relative to SubspaceAD using the same
backbone and $0.71$\,ms relative to Gaussian smoothing.

\section{Experiments}
\label{sec:experiments}

\begin{table*}[t]
\centering
\caption{Strictly Paired 5-Seed Comparison on MVTec-AD (15 Categories), VisA (12 Categories), and All 27 Categories (1/2/4-Shot, Mean $\pm$ Std, \%; $\Delta$ Computed Prior to Rounding). Both Methods Share Identical \texttt{DINOv3-L/16@512} Features and Splits. $^\dagger p < 0.05$, $^\ddagger p < 0.01$ (5-Seed Paired $t$-Test).}
\label{tab:main_5seed}
\footnotesize
\setlength{\tabcolsep}{5.0pt}
\renewcommand{\arraystretch}{1.08}
\begin{fitdatatable}\begin{tabular}{@{}clccccccccc@{}}
\toprule
\textbf{Shot} & \textbf{Method} & \multicolumn{3}{c}{\textbf{MVTec-AD (15 Categories)}} & \multicolumn{3}{c}{\textbf{VisA (12 Categories)}} & \multicolumn{3}{c}{\textbf{All 27 Categories}} \\
\cmidrule(lr){3-5} \cmidrule(lr){6-8} \cmidrule(lr){9-11}
 & & I-AUROC & P-AUROC & PRO & I-AUROC & P-AUROC & PRO & I-AUROC & PRO & Comp. \\
\midrule
\multirow{3}{*}{1-shot}
 & SubspaceAD & 96.15$\pm$1.07 & 97.28$\pm$0.06 & 94.24$\pm$0.13 & 91.42$\pm$0.74 & 97.72$\pm$0.09 & 93.62$\pm$0.38 & 94.05$\pm$0.73 & 93.96$\pm$0.18 & 95.17$\pm$0.30 \\
 & \textbf{DPLM-AD (Ours)} & \textbf{96.77$\pm$0.96} & \textbf{97.86$\pm$0.04} & \textbf{94.89$\pm$0.15} & \textbf{93.71$\pm$0.84} & \textbf{98.52$\pm$0.03} & \textbf{95.67$\pm$0.23} & \textbf{95.41$\pm$0.77} & \textbf{95.24$\pm$0.11} & \textbf{96.27$\pm$0.29} \\
 & $\Delta$ (Ours $-$ Base) & $+0.61^\dagger$ & $+0.58^\ddagger$ & $+0.66^\ddagger$ & $+2.28^\ddagger$ & $+0.80^\ddagger$ & $+2.05^\ddagger$ & $+1.35^\ddagger$ & $+1.28^\ddagger$ & $+1.10^\ddagger$ \\
\midrule
\multirow{3}{*}{2-shot}
 & SubspaceAD & 97.16$\pm$0.70 & 97.54$\pm$0.13 & 94.64$\pm$0.12 & 92.27$\pm$0.28 & 97.78$\pm$0.04 & 93.79$\pm$0.15 & 94.99$\pm$0.51 & 94.26$\pm$0.09 & 95.63$\pm$0.20 \\
 & \textbf{DPLM-AD (Ours)} & \textbf{97.58$\pm$0.78} & \textbf{98.14$\pm$0.12} & \textbf{95.33$\pm$0.14} & \textbf{94.23$\pm$0.40} & \textbf{98.59$\pm$0.03} & \textbf{95.91$\pm$0.12} & \textbf{96.10$\pm$0.57} & \textbf{95.58$\pm$0.10} & \textbf{96.67$\pm$0.21} \\
 & $\Delta$ (Ours $-$ Base) & $+0.42$ & $+0.60^\ddagger$ & $+0.69^\ddagger$ & $+1.96^\ddagger$ & $+0.81^\ddagger$ & $+2.12^\ddagger$ & $+1.11^\ddagger$ & $+1.33^\ddagger$ & $+1.04^\ddagger$ \\
\midrule
\multirow{3}{*}{4-shot}
 & SubspaceAD & 97.38$\pm$0.66 & 97.58$\pm$0.08 & 94.76$\pm$0.06 & 92.73$\pm$0.23 & 97.82$\pm$0.04 & 93.88$\pm$0.08 & 95.31$\pm$0.29 & 94.37$\pm$0.04 & 95.79$\pm$0.09 \\
 & \textbf{DPLM-AD (Ours)} & \textbf{98.06$\pm$0.71} & \textbf{98.23$\pm$0.07} & \textbf{95.46$\pm$0.11} & \textbf{94.45$\pm$0.34} & \textbf{98.60$\pm$0.03} & \textbf{95.92$\pm$0.10} & \textbf{96.45$\pm$0.34} & \textbf{95.67$\pm$0.07} & \textbf{96.84$\pm$0.13} \\
 & $\Delta$ (Ours $-$ Base) & $+0.68^\ddagger$ & $+0.65^\ddagger$ & $+0.70^\ddagger$ & $+1.71^\ddagger$ & $+0.78^\ddagger$ & $+2.04^\ddagger$ & $+1.14^\ddagger$ & $+1.30^\ddagger$ & $+1.05^\ddagger$ \\
\bottomrule
\end{tabular}\end{fitdatatable}
\end{table*}

\begin{table*}[t]
\centering
\caption{Comparison With State-of-the-Art FSAD Methods Under Published Protocols (\%). \textbf{Bold} and \underline{Underline} Indicate Highest and Second-Highest Distinct Training-Free Values per Column and Shot (`--' Denotes Unreported Metric).}
\label{tab:sota_literature}
\footnotesize
\setlength{\tabcolsep}{4.8pt}
\renewcommand{\arraystretch}{1.05}
\begin{fitdatatable}\begin{tabular}{@{}llcccccccc@{}}
\toprule
\multirow{2}{*}{\textbf{Shot}} & \multirow{2}{*}{\textbf{Method}} & \multirow{2}{*}{\textbf{Training-Free}} & \multirow{2}{*}{\textbf{Backbone \& Resolution}} & \multicolumn{3}{c}{\textbf{MVTec-AD (15)}} & \multicolumn{3}{c}{\textbf{VisA (12)}} \\
\cmidrule(lr){5-7} \cmidrule(lr){8-10}
 & & & & I-AUROC & P-AUROC & PRO & I-AUROC & P-AUROC & PRO \\
\midrule
\multirow{8}{*}{1-shot}
 & WinCLIP+~\cite{jeong2023winclip} & $\checkmark$ & CLIP-B/16+@240 & 93.1 & 95.2 & 87.1 & 83.8 & 96.4 & 85.1 \\
 & PromptAD~\cite{li2024promptad} & $\times$ & CLIP-B/16+@240 & 94.6 & 95.9 & 87.9 & 86.9 & 96.7 & 85.1 \\
 & AnomalyDINO-S~\cite{damm2025anomalydino} & $\checkmark$ & DINOv2-S/14@672 & 96.6 & 96.8 & 92.7 & 87.4 & 97.8 & 92.5 \\
 & ReMP-AD~\cite{ma2025rempad} & $\times$ (Aux.) & CLIP-L/14@518 & 94.7 & 95.6 & 90.2 & 91.3 & 97.6 & 92.2 \\
 & DCP-SFR~\cite{jiang2026dcpsfr} & $\times$ (Aux.) & CLIP-L/14@336 + DINOv2 & 97.3 & 96.9 & 92.7 & 92.0 & 98.2 & 93.3 \\
 & FastRef (AnomalyDINO+)~\cite{li2026fastref} & $\checkmark$ & DINOv2-S/14@448 & \underline{97.0} & 96.5 & -- & 84.6 & 97.2 & -- \\
 & SubspaceAD~\cite{lendering2026subspacead} & $\checkmark$ & DINOv2-G/14@672 (1.1B) & \textbf{97.1} & \underline{97.5} & \underline{93.7} & \underline{93.4} & \underline{98.2} & \underline{93.5} \\
 & \textbf{DPLM-AD (Ours)} & $\checkmark$ & \textbf{DINOv3-L/16@512 (304M)} & 96.8 & \textbf{97.9} & \textbf{94.9} & \textbf{93.7} & \textbf{98.5} & \textbf{95.7} \\
\midrule
\multirow{8}{*}{2-shot}
 & WinCLIP+~\cite{jeong2023winclip} & $\checkmark$ & CLIP-B/16+@240 & 94.4 & 96.0 & 88.4 & 84.6 & 96.8 & 86.2 \\
 & PromptAD~\cite{li2024promptad} & $\times$ & CLIP-B/16+@240 & 95.7 & 96.2 & 88.5 & 88.3 & 97.1 & 85.8 \\
 & AnomalyDINO-S~\cite{damm2025anomalydino} & $\checkmark$ & DINOv2-S/14@672 & 96.9 & 97.0 & 93.1 & 89.7 & 98.0 & 93.4 \\
 & ReMP-AD~\cite{ma2025rempad} & $\times$ (Aux.) & CLIP-L/14@518 & 95.9 & 96.2 & 91.6 & 92.9 & 97.8 & 92.6 \\
 & DCP-SFR~\cite{jiang2026dcpsfr} & $\times$ (Aux.) & CLIP-L/14@336 + DINOv2 & 97.7 & 97.1 & 93.3 & 92.5 & 98.4 & 93.8 \\
 & FastRef (AnomalyDINO+)~\cite{li2026fastref} & $\checkmark$ & DINOv2-S/14@448 & \underline{97.2} & 96.7 & -- & 84.8 & 97.3 & -- \\
 & SubspaceAD~\cite{lendering2026subspacead} & $\checkmark$ & DINOv2-G/14@672 (1.1B) & \textbf{97.6} & \underline{97.7} & \underline{94.1} & \underline{93.7} & \underline{98.3} & \underline{93.6} \\
 & \textbf{DPLM-AD (Ours)} & $\checkmark$ & \textbf{DINOv3-L/16@512 (304M)} & \textbf{97.6} & \textbf{98.1} & \textbf{95.3} & \textbf{94.2} & \textbf{98.6} & \textbf{95.9} \\
\midrule
\multirow{8}{*}{4-shot}
 & WinCLIP+~\cite{jeong2023winclip} & $\checkmark$ & CLIP-B/16+@240 & 95.2 & 96.2 & 89.0 & 87.3 & 97.2 & 87.6 \\
 & PromptAD~\cite{li2024promptad} & $\times$ & CLIP-B/16+@240 & 96.6 & 96.5 & 90.5 & 89.1 & 97.4 & 86.2 \\
 & AnomalyDINO-S~\cite{damm2025anomalydino} & $\checkmark$ & DINOv2-S/14@672 & 97.7 & 97.2 & 93.4 & 92.6 & 98.2 & \underline{94.1} \\
 & ReMP-AD~\cite{ma2025rempad} & $\times$ (Aux.) & CLIP-L/14@518 & 96.8 & 96.6 & 92.8 & 94.1 & 97.8 & 92.9 \\
 & DCP-SFR~\cite{jiang2026dcpsfr} & $\times$ (Aux.) & CLIP-L/14@336 + DINOv2 & 98.0 & 97.5 & 93.5 & 92.8 & 98.6 & 94.2 \\
 & FastRef (AnomalyDINO+)~\cite{li2026fastref} & $\checkmark$ & DINOv2-S/14@448 & 97.4 & 96.8 & -- & 87.1 & 97.9 & -- \\
 & SubspaceAD~\cite{lendering2026subspacead} & $\checkmark$ & DINOv2-G/14@672 (1.1B) & \underline{97.9} & \underline{97.8} & \underline{94.3} & \textbf{94.7} & \underline{98.4} & 93.8 \\
 & \textbf{DPLM-AD (Ours)} & $\checkmark$ & \textbf{DINOv3-L/16@512 (304M)} & \textbf{98.1} & \textbf{98.2} & \textbf{95.5} & \underline{94.4} & \textbf{98.6} & \textbf{95.9} \\
\bottomrule
\end{tabular}\end{fitdatatable}
\end{table*}

\subsection{Experimental Setup}
\label{sec:exp_setup}
We evaluate DPLM-AD across \textbf{five industrial benchmarks totaling 41 categories} ($23{,}700+$ images): \textbf{MVTec-AD}~\cite{bergmann2019mvtec} (15 categories) and \textbf{VisA}~\cite{zou2022spot} (12 categories) form the 27-category core suite; \textbf{MPDD}~\cite{jezek2021mpdd} (6 metal-part categories with 3D pose/background shifts) and \textbf{BTAD}~\cite{mishra2021vtadl} (3 surface/body products) extend evaluation to 36 structural categories; and \textbf{MVTec-LOCO}~\cite{bergmann2022beyond} (5 categories, evaluated both pooled and split into \texttt{Structural} vs.\ \texttt{Logical} anomalies) probes the boundary of positional debiasing.

Following SubspaceAD~\cite{lendering2026subspacead}, we run $k \in \{1, 2, 4\}$-shot experiments over five random seeds ($\{0, 1, 2, 3, 4\}$) with identical reference splits and augmentations (30 rotations for orientation-invariant classes; disabled for \texttt{transistor}). We report I-AUROC, I-AUPR, P-AUROC, AUPRO (PRO up to $0.30$ FPR), and $\text{Comp.} = (\text{I-AUROC} + \text{P-AUROC} + \text{PRO})/3$, testing significance via two-sided paired $t$-tests over the 5 seed macro-means and two-sided Wilcoxon signed-rank tests (\texttt{scipy.stats.wilcoxon}, excluding exact zero differences) paired across the 27 category-level 5-seed means. All defaults are fixed globally across all 41 categories: \texttt{DINOv3-L/16@512} ($P=1{,}024$), $\mathcal{L}=\{13,\dots,19\}$ ($D=1{,}024$), $s=32$, $M=12$, OAS ($\gamma_{\min}=0.01, \gamma_{\max}=0.95$), $\tau=0.95$, $r_{\max}=192$, and diffusion $(w_g=5, \tau_g=0.1, \eta=0.5, T=3)$ on one RTX 4090 GPU. Full per-category tables are in the Supplementary Material.

\subsection{Main Comparison on MVTec-AD and VisA}
\label{sec:exp_main}

\begin{table*}[t]
\centering
\caption{System Progression With Dual-Space Graph Controls (Part a) and Confounder-Free Matched-Estimator $4\times 2$ Projection $\times$ PCA Ablation (Part b) Across All 27 Categories (1-Shot, 5-Seed Mean $\pm$ Std, \texttt{DINOv3-L/16@512}, \%). In Part (a), \textit{Head (ms)} Is Post-Backbone Latency; A3, A3b, A4, and A4-sym Operate on Strictly Identical Initial Residuals $\mathbf{s}^{(0)}$. In Part (b), All Eight Cells Share Identical OAS ($\tau=0.95, r_{\max}=192$) and Gaussian ($\sigma=4$).}
\label{tab:ablation_matched}
\footnotesize
\setlength{\tabcolsep}{4.0pt}
\renewcommand{\arraystretch}{1.06}
\begin{fitdatatable}\begin{tabular}{@{}llccccccccc@{}}
\toprule
\textbf{Study / Variant} & \textbf{Projection ($s$)} & \textbf{$M$} & \textbf{Post-Proc.} & \textbf{I-AUROC} & \textbf{I-AUPR} & \textbf{P-AUROC} & \textbf{PRO} & \textbf{Comp.} & \textbf{Head / $\Delta_{\text{syn}}$ (I)} & \textbf{Mem / $p$-val} \\
\midrule
\multicolumn{11}{l}{\textit{(a) End-to-End Progression \& Paired Diffusion Controls on Identical $\mathbf{s}^{(0)}$ (A0 uses SubspaceAD $\tau=0.99$; A1--A5 use OAS, $\tau=0.95, r_{\max}=192$)}} \\
A0 (SubspaceAD) & None ($0$) & $1$ & Gaussian & 94.05$\pm$0.73 & 95.08$\pm$0.43 & 97.48$\pm$0.04 & 93.96$\pm$0.18 & 95.17$\pm$0.30 & 5.32\,ms & 1.67\,MB \\
A1 (+ Debias only) & LowFreq ($32$) & $1$ & Gaussian & 94.14$\pm$0.80 & 95.11$\pm$0.46 & 97.42$\pm$0.04 & 94.00$\pm$0.18 & 95.18$\pm$0.33 & 3.85\,ms & \textbf{0.71\,MB} \\
A2 (+ Local only) & None ($0$) & $12$ & Gaussian & 91.09$\pm$1.02 & 91.93$\pm$0.75 & 97.39$\pm$0.18 & 92.73$\pm$0.70 & 93.74$\pm$0.60 & \textbf{3.69\,ms} & 3.08\,MB \\
A3 (+ Debias + Local) & LowFreq ($32$) & $12$ & Gaussian & 94.91$\pm$0.76 & 95.68$\pm$0.65 & 97.83$\pm$0.03 & 94.70$\pm$0.10 & 95.81$\pm$0.28 & 5.30\,ms & 6.91\,MB \\
A3b (+ Deb-Graph Diff.) & LowFreq ($32$) & $12$ & Deb-Graph & 95.16$\pm$0.75 & 96.00$\pm$0.60 & 98.11$\pm$0.02 & 95.16$\pm$0.11 & 96.14$\pm$0.28 & 5.99\,ms & 6.91\,MB \\
\textbf{A4 (DPLM-AD Full)} & \textbf{LowFreq ($32$)} & \textbf{$12$} & \textbf{Raw-Graph} & 95.41$\pm$0.77 & 96.24$\pm$0.57 & 98.15$\pm$0.02 & 95.24$\pm$0.11 & 96.27$\pm$0.29 & 6.01\,ms & 6.91\,MB \\
A4-sym (Valid-Mask) & LowFreq ($32$) & $12$ & Raw-Graph (Sym) & \textbf{95.46$\pm$0.79} & \textbf{96.29$\pm$0.57} & 98.16$\pm$0.02 & \textbf{95.25$\pm$0.11} & \textbf{96.29$\pm$0.30} & 6.03\,ms & 6.91\,MB \\
A5 (DPLM-AD Light) & LowFreq ($16$) & $8$ & Raw-Graph & 95.25$\pm$0.79 & 96.06$\pm$0.49 & \textbf{98.18$\pm$0.03} & 95.23$\pm$0.17 & 96.22$\pm$0.31 & 5.52\,ms & 4.31\,MB \\
\midrule
\multicolumn{11}{l}{\textit{(b) Strictly Matched-Estimator $4\times 2$ Ablation (Fixed OAS, $\tau=0.95, r_{\max}=192$, Gaussian; Rightmost Columns Show $2\times 2$ Synergy $\Delta_{\text{syn}}$ vs.\ No-Proj)}} \\
No-Proj & None ($0$) & $1$ / $12$ & Gaussian & 93.96 / 91.09 & 95.29 / 91.93 & 97.05 / 97.39 & 93.48 / 92.73 & 94.83 / 93.74 & Baseline ($-2.87$) & -- \\
Rand-Proj & Random ($32$) & $1$ / $12$ & Gaussian & 93.96 / 91.12 & \textbf{95.32} / 91.92 & 97.04 / 97.38 & 93.47 / 92.72 & 94.82 / 93.74 & $+0.03\pm 0.20$ & $p=0.767$ \\
INSID3-Noise~\cite{cuttano2026insid3} & WhiteNoise ($32$) & $1$ / $12$ & Gaussian & \textbf{94.18} / 94.83 & 95.12 / \textbf{95.73} & \textbf{97.42} / \textbf{97.86} & 93.94 / 94.65 & \textbf{95.18} / 95.78 & $+3.53\pm 1.36$ & $p=0.0044$ \\
\textbf{Ours-LowFreq} & \textbf{LowFreq ($32$)} & $1$ / $12$ & Gaussian & 94.14 / \textbf{94.91} & 95.11 / 95.68 & \textbf{97.42} / 97.83 & \textbf{94.00} / \textbf{94.70} & \textbf{95.18} / \textbf{95.81} & \textbf{$+3.65\pm 1.35$} & \textbf{$p=0.0037$} \\
\bottomrule
\end{tabular}\end{fitdatatable}
\end{table*}

\begin{figure*}[t]
\centering
\includegraphics[width=\textwidth]{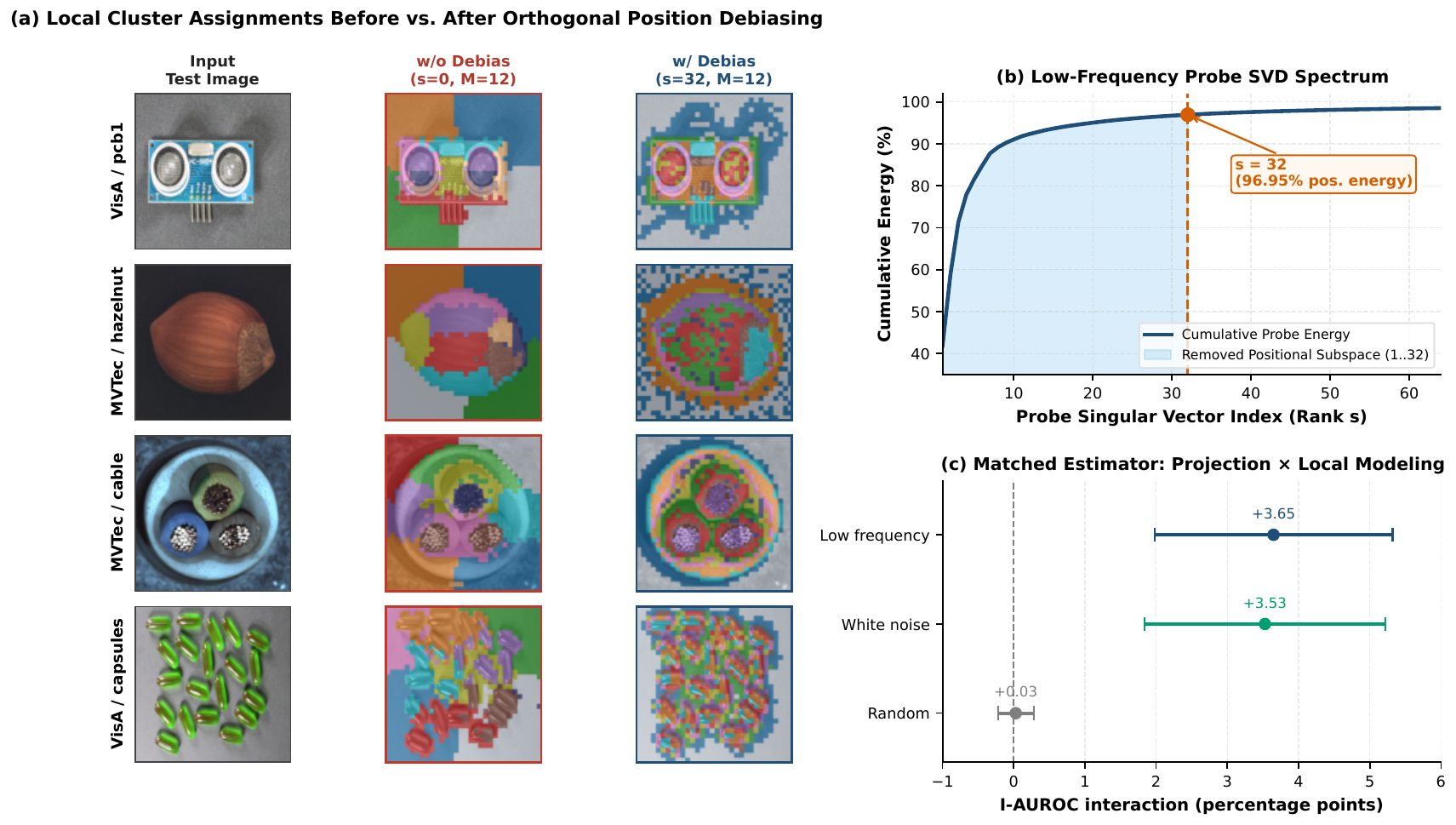}
\caption{Visual and statistical verification of positional-manifold synergy. (a) $M=12$ cluster assignments on pose-shifted test images before ($s=0$) vs.\ after ($s=32$) orthogonal debiasing: raw clusters fracture into spatial quadrants, whereas debiased clusters track physical parts. (b) Cumulative singular energy of the 12 low-frequency probes ($s=32$ captures $96.95\%$). (c) Confounder-free matched-estimator $2\times 2$ I-AUROC interaction $\Delta_{\text{syn}}$ (mean and paired 95\% $t$-interval over 5 reference seeds across 27 categories).}
\label{fig:mechanism}
\end{figure*}

Table~\ref{tab:main_5seed} reports the strictly paired 5-seed comparison under identical \texttt{DINOv3-L/16@512} features, while Table~\ref{tab:sota_literature} compares against published FSAD methods:
\begin{enumerate}
    \item \textbf{Consistent Gains Across Every Shot Regime:} Under shared features (Table~\ref{tab:main_5seed}), DPLM-AD lifts All-27 I-AUROC by $+1.35\% / +1.11\% / +1.14\%$, PRO by $+1.28\% / +1.33\% / +1.30\%$, and Composite by $+1.10\% / +1.04\% / +1.05\%$ at 1/2/4-shot (5-seed paired $t$-test $p < 10^{-3}$). Across the 27 category-level 5-seed means, two-sided Wilcoxon tests confirm strong significance across every shot and metric: $p < 4.3\times 10^{-4}$ at 1/2-shot ($p = 4.5\times 10^{-8}$ for 1-shot Comp., $26/27$ wins), and at 4-shot $p = 1.47\times 10^{-3}$ for I-AUROC/I-AUPR ($18/27$ wins) and $p < 7\times 10^{-6}$ for P-AUROC/PRO/Comp.\ ($27/27$ P-AUROC wins; exact per-metric $p$-values in Supplementary Section~S4-B; paired over all 135 category--seed runs, every metric gives $p < 10^{-9}$). On 1-shot VisA, DPLM-AD wins \textbf{$12/12$ categories} across all metrics ($+2.28\%$ I-AUROC, $+2.05\%$ PRO).
    \item \textbf{$4\times$ Localization Efficiency and Multi-Part Scaling:} On the shared \texttt{DINOv3-L/16} backbone, 1-shot DPLM-AD surpasses 4-shot SubspaceAD in PRO ($95.24\%$ vs.\ $94.37\%$, $+0.870$ points, 95\% CI $[0.756, 0.985]$, Fig.~\ref{fig:teaser}(b)) and Composite ($96.27\%$ vs.\ $95.79\%$). On multi-part VisA circuit boards (\texttt{pcb1}--\texttt{pcb4}), our I-AUROC margin over global PCA \emph{widens} with reference shots ($+2.55\% \to +3.51\% \to +4.04\%$; PRO $+4.05\% \to +4.71\% \to +5.07\%$).
    \item \textbf{Comparison With Published Heavyweight SOTA:} Even against published SubspaceAD using a $3.6\times$ larger 1.1B \texttt{DINOv2-G/14@672} backbone ($2.25\times$ denser token grid), 304M DPLM-AD@512 achieves the \textbf{highest P-AUROC and PRO across both datasets and all three shots} ($+1.2\%$ MVTec and $+2.2\%$ VisA PRO at 1-shot), matches or exceeds \texttt{DINOv2-G/14@672} I-AUROC on 2/4-shot MVTec-AD and 1/2-shot VisA, and outperforms auxiliary-trained DCP-SFR~\cite{jiang2026dcpsfr} across all VisA metrics.
\end{enumerate}

\subsection{Mechanism Verification and Empirical Phase Diagram}
\label{sec:exp_ablation}

\begin{figure*}[t]
\centering
\includegraphics[width=\textwidth]{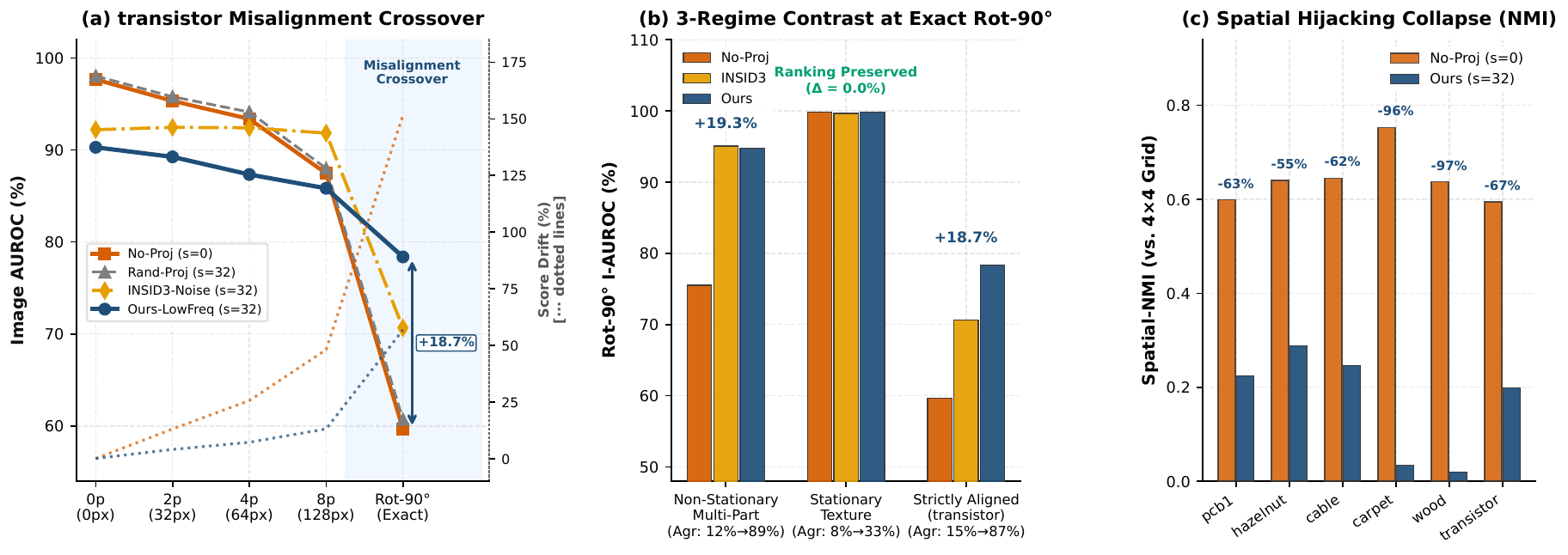}
\caption{Controlled three-regime pose counterfactual study on six representative categories ($M=12$, 1-shot, \texttt{seed=0}). (a) On strictly aligned \texttt{transistor}, \texttt{No-Proj} exploits coordinate priors at zero shift (\texttt{0p}) but incurs severe normal score drift ($+151.5\%$) and diagnostic FPR saturation ($100\%$ at \texttt{4p}), crossing over to a $+18.7\%$ I-AUROC win for \texttt{Ours-LowFreq} at exact $90^\circ$ rotation. (b) At exact \texttt{Rot-90$^\circ$}, debiasing lifts non-stationary multi-part I-AUROC by $+19.3\%$ (physical cluster agreement $12\% \to 89\%$), while stationary texture ranking remains preserved ($\Delta = 0.0\%$). (c) \texttt{Spatial-NMI} against a $4\times 4$ coordinate grid drops by $55\%\text{--}97\%$.}
\label{fig:pose_counterfactual}
\end{figure*}

\begin{figure*}[t]
\centering
\includegraphics[width=\textwidth]{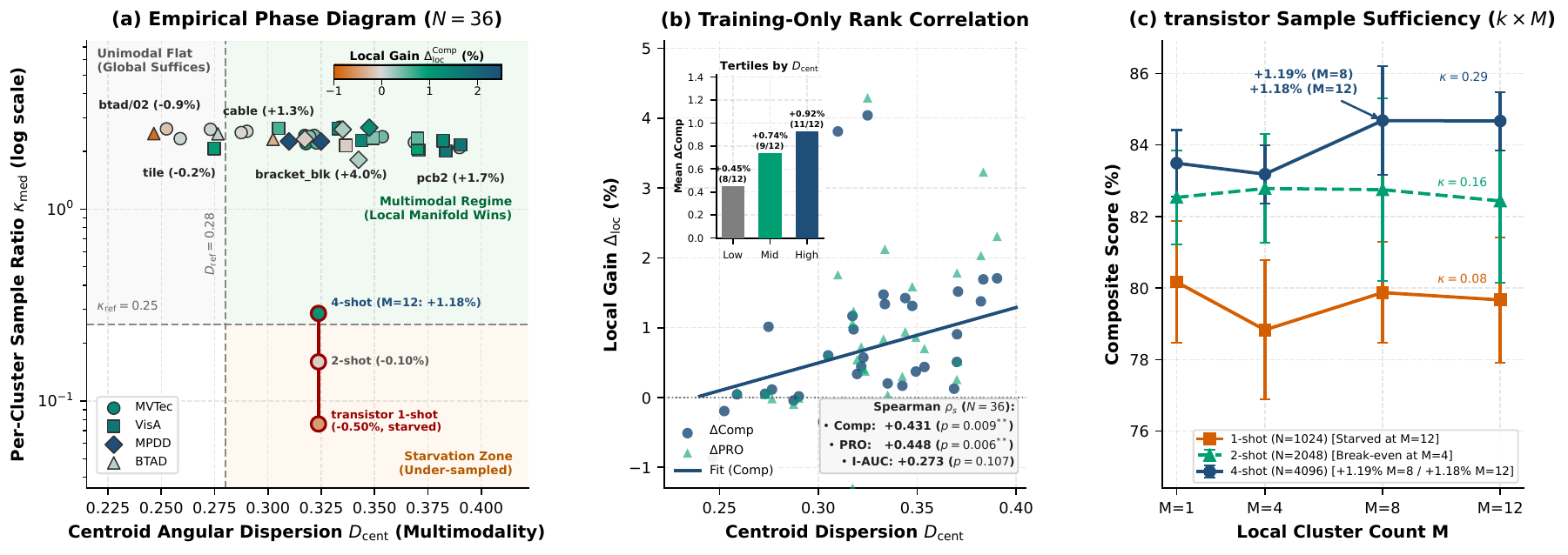}
\caption{Zero-test-leakage empirical phase diagram across 36 structural categories (1-shot, 5 seeds, matched OAS + Gaussian). (a) Phase space of training-only centroid angular dispersion $D_{\text{cent}}$ and median sample-to-dimension ratio $\kappa_{\text{med}} = \mathrm{median}_m(n_m / d)$ ($d = D - s = 992$), with exploratory reference lines $D_{\text{ref}}=0.28, \kappa_{\text{ref}}=0.25$ and the $1\to 2\to 4$-shot $M=12$ trajectory of \texttt{transistor} (at 1-shot $D_{\text{cent}}=0.324$). (b) Spearman rank correlations across $N=36$ categories ($\rho_s=+0.431$ Comp., $+0.448$ PRO, $+0.273$ I-AUROC) and tertile gains (inset). (c) Unrotated \texttt{transistor} scaling ($k \times M$), resolving 1-shot starvation ($\kappa=0.08$) into a $+1.19\%$ ($M=8$) / $+1.18\%$ ($M=12$) 4-shot win ($\kappa \ge 0.29$).}
\label{fig:phase_diagram}
\end{figure*}

\subsubsection{Matched-Estimator Synergy, Equal-Budget Probes, and Dual-Space Graph Control}
Table~\ref{tab:ablation_matched}(b) holds OAS, rank selection, and Gaussian smoothing \emph{strictly identical} across four projections crossed with $M \in \{1, 12\}$. Defining $g_j(Q) = \text{I-AUROC}_j(Q, M=12) - \text{I-AUROC}_j(Q, M=1)$ on seed $j$, the paired $2\times 2$ interaction $\Delta_{\text{syn}}(Q) = \frac{1}{5}\sum_{j=1}^5 [g_j(Q) - g_j(Q_0)]$ yields four key conclusions (Fig.~\ref{fig:mechanism}; full tables in Supplementary Section~S3):
\begin{itemize}
    \item \emph{Falsification of random projection:} \texttt{Rand-Proj} ($s=32$, Grassmann mean $\cos\theta = 0.151$) leaves the $M=1\to 12$ collapse intact ($\Delta_{\text{syn}} = +0.03 \pm 0.20\%$, $p=0.767$).
    \item \emph{Strong interaction with category heterogeneity:} \texttt{Ours-LowFreq} reverses $-2.87\%$ to $+0.77\%$, producing $\Delta_{\text{syn}} = +3.65 \pm 1.35\%$ I-AUROC (95\% CI $[+1.98, +5.32]$, $p=0.00374$, positive on all 5 seeds) and $+1.72 \pm 0.62\%$ Composite ($p=0.0035$). Across the 27 category 5-seed means, the interaction is heterogeneous as predicted by geometry: positive on $19/27$ categories in I-AUROC ($23/27$ in Composite), averaging $+1.40\%$ on MVTec-AD vs.\ $+6.46\%$ on multi-part VisA. (The 5-seed CI reflects reference-sampling variation; cross-category generalization is tested on 36 categories below.)
    \item \emph{Probe-universal subspace removal \& real-image evidence:} Under an equal 12-probe budget across 5 paired data/probe seeds (Supplementary Table~S3), \texttt{LowFreq-12} ($+3.74\%$ synergy), purely deterministic \texttt{Analytic-9} ($+3.64\%$), \texttt{WhiteNoise-12} ($+3.62\%$), and single-image \texttt{WhiteNoise-1} ($+3.68\%$) perform nearly identically (paired $M=12$ \texttt{LowFreq-12} $-$ \texttt{WhiteNoise-12} is $+0.12\%$ I-AUROC, $p=0.333$, and $+0.08\%$ Comp., $p=0.170$ with diffusion). Thus, the core gain comes from removing the shared positional subspace rather than a probe-specific artifact, while 12-probe averaging roughly halves fixed-data probe-seed standard deviation relative to single-noise ($0.043\%/0.054\%$ vs.\ $0.088\%$ I-AUROC std). Critically, on unseen real normal test images (27 categories), the 32D probe subspace $B$ captures $66.2 \pm 2.6\%$ of uncentered energy, $81.2 \pm 3.1\%$ of centered variance ($31\times$ random's $2.62\%$; vs.\ $79.0\%$ for \texttt{WhiteNoise-12}), and $88.8 \pm 5.0\%$ of pure mean-grid spatial variance ($R^2_{\text{pos}} = 0.992$ from 32D $B$ alone vs.\ $0.997$ raw 1024D).
    \item \emph{Direct dual-space diffusion control (A3 vs.\ A3b vs.\ A4 vs.\ A4-sym):} Operating on \emph{strictly identical} initial residuals $\mathbf{s}^{(0)}$ (Table~\ref{tab:ablation_matched}(a) and Supplementary Table~S2), replacing Gaussian (A3) with debiased-feature graph diffusion (A3b) adds $+0.252\%$ I-AUROC and $+0.456\%$ PRO; switching the graph guidance from debiased $\hat{F}$ back to raw $F$ (A3b $\to$ A4) independently adds another \textbf{$+0.243\%$ I-AUROC} (95\% CI $[+0.158, +0.328]$, paired $t$ $p=0.00134$, Wilcoxon $p=2.5\times 10^{-4}$, $19/27$ wins), \textbf{$+0.082\%$ PRO} ($p=5.1\times 10^{-5}$, $20/27$ wins), and \textbf{$+0.123\%$ Composite} ($p=5.0\times 10^{-4}$, Wilcoxon $p=8.0\times 10^{-6}$, \textbf{$24/27$ wins}), with largest gains on \texttt{transistor} ($+1.41\%$ I / $+0.45\%$ PRO) and \texttt{pcb2}--\texttt{pcb4} (e.g., $+3.40\%$ single-image PRO on \texttt{pcb2/018.JPG} in Fig.~\ref{fig:qualitative}). Enforcing exact global symmetry via valid-neighbor masking (\texttt{A4-sym}) adds another $+0.055\%$ I-AUROC ($p=0.015$) and $+0.024\%$ Comp.\ ($22/27$ wins, $96.29\%$).
\end{itemize}

\subsubsection{Controlled Three-Regime Pose Counterfactuals}
\label{sec:exp_counterfactual}

\begin{table*}[t]
\centering
\caption{Geometric Representation Study Across All 27 Categories (1-Shot, \texttt{seed=0}, \texttt{DINOv3-L/16@512} With Graph Diffusion, \%). \textit{Post-BB (ms)} Is Batched GPU Subspace Scoring Plus Shared Diffusion/Upsampling Latency (SubspaceAD Uses CPU PCA + Gaussian). Matched $(s=48, M=8)$ Rows Directly Isolate Euclidean vs.\ Spherical Geometry.}
\label{tab:geometry_study}
\footnotesize
\setlength{\tabcolsep}{5.2pt}
\renewcommand{\arraystretch}{1.06}
\begin{fitdatatable}\begin{tabular}{@{}llccccccc@{}}
\toprule
\textbf{Geometry Space} & \textbf{Variant ($s, M$)} & \textbf{All-27 I-AUROC} & \textbf{All-27 P-AUROC} & \textbf{All-27 PRO} & \textbf{All-27 Comp.} & \texttt{screw} (I) & \texttt{pcb3} (I) & \textbf{Post-BB (ms)} \\
\midrule
Ambient Euclidean & SubspaceAD ($0, 1$) & 94.12 & 97.54 & 94.14 & 95.27 & 86.25 & 88.93 & 7.45 \\
\midrule
Unit Sphere ($\Sp^{d-1}$) & Sphere-Local ($48, 8$) [Matched] & 94.85 & 97.97 & 94.39 & 95.74 & 75.43 & 83.81 & 6.74 \\
Unit Sphere ($\Sp^{d-1}$) & Sphere-Hier ($48, 1+8$) & 95.21 & 97.95 & 94.67 & 95.94 & 75.88 & 85.27 & 4.83 \\
\midrule
Product ($\R_+ \times \Sp^{d-1}$) & Unwarped-Hier ($48, 1+8$) & 95.50 & 97.91 & 94.81 & 96.08 & 82.84 & 90.65 & 3.58 \\
Warped Product ($\R_+ \times_r \Sp^{d-1}$) & Warped-Hier ($16, 1+8$) & 95.82 & 97.93 & 95.00 & 96.25 & \textbf{89.98} & \textbf{93.94} & 3.71 \\
\midrule
Local Euclidean ($\R^{D-s}$) & Euclid-Local ($48, 8$) [Matched] & 95.93 & 98.15 & 95.26 & 96.45 & 84.87 & 90.53 & \textbf{3.23} \\
Local Euclidean ($\R^{D-s}$) & Euclid-Local ($16, 8$) & 95.99 & 98.17 & 95.28 & 96.48 & 86.99 & 86.44 & 3.27 \\
\textbf{Local Euclidean ($\R^{D-s}$)} & \textbf{DPLM-AD ($32, 12$)} & \textbf{96.01} & \textbf{98.18} & \textbf{95.32} & \textbf{96.50} & 87.15 & 91.10 & 3.45 \\
\bottomrule
\end{tabular}\end{fitdatatable}
\end{table*}

\begin{table*}[t]
\centering
\caption{Cross-Backbone Generalization and Efficiency Across All 27 Categories (1-Shot, 5-Seed Mean $\pm$ Std, \%). \textit{Inf (ms)} Includes Backbone + Head; \textit{Mem (MB)} Reports Per-Category FP32 State (DPLM-AD Halves to $3.43$\,MB in Lossless FP16).}
\label{tab:backbones_5seed}
\footnotesize
\setlength{\tabcolsep}{4.6pt}
\renewcommand{\arraystretch}{1.06}
\begin{fitdatatable}\begin{tabular}{@{}llcccccccc@{}}
\toprule
\textbf{Backbone (\#Params, Res.)} & \textbf{Method} & \textbf{MVTec Comp.} & \textbf{VisA Comp.} & \textbf{All-27 I-AUROC} & \textbf{All-27 P-AUROC} & \textbf{All-27 PRO} & \textbf{All-27 Comp.} & \textbf{Inf (ms)} & \textbf{Mem (MB)} \\
\midrule
\multirow{2}{*}{\texttt{DINOv2-L/14} (304M, 518)}
 & SubspaceAD & 95.63$\pm$0.37 & 94.45$\pm$0.33 & 93.56$\pm$0.51 & 97.54$\pm$0.05 & 94.22$\pm$0.20 & 95.11$\pm$0.24 & 55.11 & 7.12 \\
 & \textbf{DPLM-AD} & \textbf{96.05$\pm$0.29} & \textbf{94.82$\pm$0.24} & \textbf{94.31$\pm$0.42} & \textbf{97.67$\pm$0.07} & \textbf{94.54$\pm$0.19} & \textbf{95.50$\pm$0.21} & \textbf{52.86} & 7.83 \\
\midrule
\multirow{2}{*}{\texttt{DINOv2-G/14} (1.1B, 518)}
 & SubspaceAD & 95.86$\pm$0.41 & 95.05$\pm$0.29 & 94.26$\pm$0.50 & \textbf{97.81$\pm$0.03} & 94.43$\pm$0.17 & 95.50$\pm$0.22 & 161.61 & 17.03 \\
 & \textbf{DPLM-AD} & \textbf{96.17$\pm$0.38} & \textbf{95.25$\pm$0.30} & \textbf{94.94$\pm$0.51} & 97.78$\pm$0.04 & \textbf{94.56$\pm$0.17} & \textbf{95.76$\pm$0.23} & \textbf{156.68} & \textbf{12.09} \\
\midrule
\multirow{2}{*}{\texttt{DINOv3-L/16} (304M, 512)}
 & SubspaceAD & 95.89$\pm$0.36 & 94.26$\pm$0.38 & 94.05$\pm$0.73 & 97.48$\pm$0.04 & 93.96$\pm$0.18 & 95.17$\pm$0.30 & \textbf{40.41} & \textbf{1.67} \\
 & \textbf{DPLM-AD} & \textbf{96.51$\pm$0.34} & \textbf{95.97$\pm$0.36} & \textbf{95.41$\pm$0.77} & \textbf{98.15$\pm$0.02} & \textbf{95.24$\pm$0.11} & \textbf{96.27$\pm$0.29} & 41.10 & 6.91 \\
\bottomrule
\end{tabular}\end{fitdatatable}
\end{table*}

To test Remark~\ref{rem:three_regimes}, we evaluate $M=12$ models on six representative categories (1-shot, \texttt{seed=0})—non-stationary multi-part (\texttt{pcb1}, \texttt{hazelnut}, \texttt{cable}), stationary textures (\texttt{carpet}, \texttt{wood}), and strictly aligned \texttt{transistor}—under horizontal shifts $\Delta \in \{0, 2, 4, 8\}$ patches and exact border-free $90^\circ$ rotation (\texttt{Rot-90$^\circ$}), tracking cluster agreement, normal score drift, test I-AUROC, and a diagnostic normal FPR evaluated at the fixed threshold giving $95\%$ TPR on unshifted test anomalies (Fig.~\ref{fig:pose_counterfactual}):
\begin{itemize}
    \item \emph{Spatial hijacking collapse:} \texttt{Spatial-NMI} with a $4\times 4$ coordinate grid drops by $55\%\text{--}97\%$ after debiasing (e.g., \texttt{carpet} $0.753 \to 0.033$, \texttt{wood} $0.637 \to 0.020$, \texttt{pcb1} $0.598 \to 0.224$; Fig.~\ref{fig:pose_counterfactual}(c)), whereas \texttt{Rand-Proj} stays unchanged ($0.751 / 0.676 / 0.595$).
    \item \emph{Three-regime divergence:} Under exact \texttt{Rot-90$^\circ$} (Fig.~\ref{fig:pose_counterfactual}(b)), \texttt{No-Proj} physical agreement falls to $11.9\%$ on multi-part objects ($75.52\%$ I-AUROC), whereas \texttt{Ours-LowFreq} maintains $89.0\%$ agreement and reaches \textbf{$94.77\%$ ($+19.25\%$)}. On stationary textures, ranking accuracy (I-AUROC) remains preserved at $99.82\%$ ($\Delta=0.0\%$) despite \texttt{No-Proj} quadrant agreement dropping to $7.6\%$; however, absolute calibration is not strictly invariant on anisotropic grain—on \texttt{wood}, fixed-threshold diagnostic FPR rises from $5.26\%$ to $10.53\%$ ($1/19 \to 2/19$ normal images) for both \texttt{No-Proj} and \texttt{Ours-LowFreq}. On strictly aligned \texttt{transistor} (Fig.~\ref{fig:pose_counterfactual}(a)), \texttt{No-Proj} leads at $\Delta=0$ ($97.67\%$ vs.\ $90.29\%$), but its diagnostic FPR hits $100\%$ by $64$\,px shift and its normal score inflates $+151.5\%$ at \texttt{Rot-90$^\circ$} (I-AUROC $59.67\%$), crossing over to a \textbf{$+18.71\%$ win} for \texttt{Ours-LowFreq} ($78.38\%$, $87.4\%$ agreement).
\end{itemize}

\subsubsection{36-Category Exploratory Phase Diagram and LODO Check}
\label{sec:exp_phase}

\begin{table*}[t]
\centering
\caption{Diffusion Sensitivity (All 27 Categories, 1-Shot, \texttt{seed=0}) and Lossless FP16 Compression (5-Seed All-27 Mean, \%). \textit{Cost} Is Diffusion + Upsampling Latency (a) or Minimal Per-Category State Excluding Shared $B_\perp$ (b).}
\label{tab:diff_fp16}
\footnotesize
\setlength{\tabcolsep}{4.0pt}
\renewcommand{\arraystretch}{1.06}
\begin{fitdatatable}\begin{tabular}{@{}lcccc@{\hspace{1.4em}}lcccc@{}}
\toprule
\multicolumn{5}{@{}l}{\textbf{(a) Graph Diffusion Sensitivity ($w_g=5, \tau_g=0.1, \eta=0.5, T=3$ def.)}} & \multicolumn{5}{l@{}}{\textbf{(b) FP16 Compression (+$1.94$\,MB Shared $B_\perp$)}} \\
\cmidrule(r{1.2em}){1-5} \cmidrule(l){6-10}
\textbf{Setting / Sweep} & \textbf{I-AUROC} & \textbf{PRO} & \textbf{Comp.} & \textbf{Cost} & \textbf{Model / Precision} & \textbf{I-AUROC} & \textbf{PRO} & \textbf{Comp.} & \textbf{Storage} \\
\midrule
$T = 0$ (No Diff.) & 95.53 & 94.77 & 96.05 & 1.76\,ms & DPLM-AD ($M=12$) FP32 & 95.4066 & 95.2402 & 96.2671 & 6.85\,MB \\
$T = 1$ / $T = 2$ & 95.95 / 96.01 & 95.25 / 95.31 & 96.44 / 96.50 & 2.79 / 2.82\,ms & \textbf{DPLM-AD ($M=12$) FP16} & \textbf{95.4068} & \textbf{95.2402} & \textbf{96.2672} & \textbf{3.43\,MB} \\
$T = 3$ (Def.) / $T = 7$ & 96.01 / 96.01 & 95.32 / 95.33 & 96.50 / 96.51 & 2.82 / 2.91\,ms & DPLM-AD ($M=8$) FP32 & 95.2520 & 95.2255 & 96.2197 & 4.27\,MB \\
$(w_g,\tau_g)=(3,0.1)$ / $(7,0.1)$ & 95.90 / 96.10 & 95.19 / 95.35 & 96.39 / 96.55 & 2.46 / 3.37\,ms & \textbf{DPLM-AD ($M=8$) FP16} & \textbf{95.2523} & \textbf{95.2255} & \textbf{96.2198} & \textbf{2.14\,MB} \\
$(w_g,\tau_g)=(5,0.05)$ / $(5,0.2)$ & 95.94 / 95.74 & 95.13 / 95.32 & 96.37 / 96.44 & 2.81 / 2.87\,ms & \multicolumn{5}{l@{}}{\textit{Max 5-seed macro $|\Delta| \le 0.0002\%$ ($2.00\times$ lossless compression)}} \\
\bottomrule
\end{tabular}\end{fitdatatable}
\end{table*}

\begin{figure*}[t]
\centering
\includegraphics[width=\textwidth]{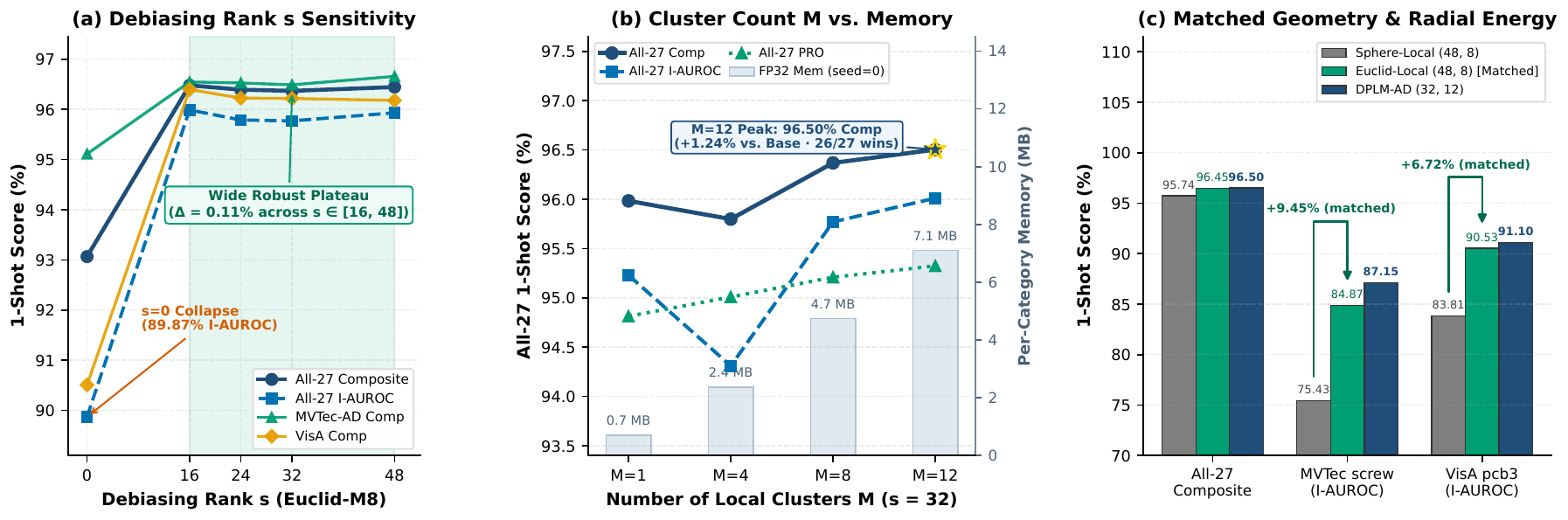}
\caption{Hyperparameter sensitivity and matched geometric ablation on All 27 categories (1-shot, \texttt{seed=0}). (a) Flat plateau across $s \in [16, 48]$ at $M=8$ ($\Delta = 0.11\%$) vs.\ $s=0$ collapse. (b) Cluster count $M \in \{1, 4, 8, 12\}$ vs.\ \texttt{seed=0} FP32 state at $s=32$ ($M=12$ peaks at $96.50\%$ Comp., $26/27$ wins; $3.43$\,MB in 5-seed FP16). (c) Strictly matched $(48, 8)$ spherical vs.\ local Euclidean comparison ($+9.45\%$ on \texttt{screw}, $+6.72\%$ on \texttt{pcb3}) alongside full DPLM-AD $(32, 12)$.}
\label{fig:sensitivity}
\end{figure*}

Across all \textbf{36 structural categories} (MVTec-15, VisA-12, MPDD-6, BTAD-3; 1-shot, 5 seeds, matched OAS + Gaussian), debiased $M=12$ improves over $M=1$ by $+1.09\%$ I-AUROC ($90.55\% \to 91.64\%$), $+0.64\%$ PRO, and $+0.70\%$ Composite ($28/36$ positive), led by multi-pose MPDD ($+3.18\%$ I-AUROC) and VisA ($+1.47\%$).

From debiased normal training references alone (zero test leakage), we compute centroid angular dispersion $D_{\text{cent}} = \frac{2}{M(M-1)}\sum_{m<l}(1 - \langle \mathbf{c}_m, \mathbf{c}_l \rangle)$ and median per-cluster sample-to-dimension ratio $\kappa_{\text{med}} = \mathrm{median}_m(n_m / d)$ ($d = D - s = 992$; Fig.~\ref{fig:phase_diagram}):
\begin{itemize}
    \item \emph{Empirical rank correlation:} Across $N=36$ categories, $D_{\text{cent}}$ correlates significantly with local gain in \textbf{PRO ($\rho_s = +0.448, p=0.0061$)} and \textbf{Composite ($\rho_s = +0.431, p=0.0087$)}, with a weaker positive trend in I-AUROC ($\rho_s = +0.274, p=0.107$; Fig.~\ref{fig:phase_diagram}(b)). Equal tertiles by $D_{\text{cent}}$ increase monotonically: Low ($[0.247, 0.317]$) $+0.45\%$ Comp.\ ($+0.04\%$ PRO), Middle $+0.74\%$, High ($[0.344, 0.390]$) $+0.92\%$ Comp.\ ($+1.09\%$ PRO, $11/12$ wins).
    \item \emph{Sample starvation on unrotated \texttt{transistor}:} Without rotation augmentation ($N=1{,}024$), 1-shot $M=12$ suffers $\kappa_{\text{med}} = 0.08$ ($-0.50\%$ Comp.\ vs.\ $M=1$). Increasing shots $k=1 \to 2 \to 4$ lifts $\kappa_{\text{med}}$ ($0.08 \to 0.16 \to 0.29$), converting starvation into a $+1.19\%$ ($M=8$) / $+1.18\%$ ($M=12$) local win (Fig.~\ref{fig:phase_diagram}(c)).
    \item \emph{Scope as an exploratory diagnostic:} Because the 35 rotated categories all sit at $\kappa_{\text{med}} \in [1.81, 2.67]$, the reference lines ($D_{\text{ref}}=0.28, \kappa_{\text{ref}}=0.25$) represent an empirical trend rather than a sharp universal decision rule. Applying them directly yields $27/36$ Composite sign accuracy vs.\ $28/36$ for always selecting $M=12$, and Leave-One-Dataset-Out (LODO) validation (Supplementary Table~S5) yields $+0.696\%$ OOF mean Composite gain ($28/36$ hits) vs.\ $+0.703\%$ ($28/36$) for always-local—mitigating extreme unimodal (\texttt{btad}: $-0.37\% \to -0.08\%$) and starved (\texttt{transistor}) outliers while confirming that fixed $M=12$ is already robust overall.
\end{itemize}

\subsection{Geometry, Backbones, Sensitivity, and FP16 Storage}
\label{sec:exp_geom}

\begin{table*}[t]
\centering
\caption{Extended Structural Benchmarks (MPDD, BTAD, All-36), Split MVTec-LOCO (\texttt{Structural} vs.\ \texttt{Logical}), and Test-Time Perturbations (\texttt{DINOv3-L/16@512}, \%). Parts (a)--(b) Report 5-Seed Paired Means ($^\dagger p<0.05, ^\ddagger p<0.01$).}
\label{tab:extra_robust}
\footnotesize
\setlength{\tabcolsep}{4.5pt}
\renewcommand{\arraystretch}{1.05}
\begin{fitdatatable}\begin{tabular}{@{}llcccc@{}}
\toprule
\textbf{Benchmark / Condition} & \textbf{Regime} & \textbf{SubspaceAD (I / P / PRO / Comp)} & \textbf{DPLM-AD (I / P / PRO / Comp)} & \textbf{$\Delta$ I-AUROC} & \textbf{$\Delta$ Comp.} \\
\midrule
\multicolumn{6}{l}{\textit{(a) Extended Structural Industrial Benchmarks (5 Seeds; All-36 Combines MVTec-15 + VisA-12 + MPDD-6 + BTAD-3)}} \\
\multirow{3}{*}{MPDD~\cite{jezek2021mpdd} (6 Metal-Part Cats)}
 & 1-shot & 70.44 / 97.07 / 91.70 / 86.40 & \textbf{74.22} / \textbf{97.59} / \textbf{93.11} / \textbf{88.31} & $+3.78^\dagger$ & $+1.91^\ddagger$ \\
 & 2-shot & 75.72 / 97.44 / 93.26 / 88.81 & \textbf{80.75} / \textbf{97.94} / \textbf{94.66} / \textbf{91.12} & $+5.03^\ddagger$ & $+2.31^\ddagger$ \\
 & 4-shot & 80.69 / 97.68 / 94.09 / 90.82 & \textbf{85.06} / \textbf{98.09} / \textbf{95.31} / \textbf{92.82} & $+4.37^\ddagger$ & $+2.00^\ddagger$ \\
\cmidrule(lr){1-6}
BTAD~\cite{mishra2021vtadl} (3 Unimodal Cats) & 1 / 4-shot & \textbf{97.80}/\textbf{94.89} \;\big|\; \textbf{97.49}/\textbf{94.72} & 97.68/94.35 \;\big|\; 97.39/94.08 & $-0.12$ / $-0.10$ & $-0.54^\ddagger$ / $-0.64^\ddagger$ \\
\cmidrule(lr){1-6}
\multirow{3}{*}{\textbf{All 36 Structural Categories}}
 & \textbf{1-shot} & 90.43 / 97.50 / 93.12 / 93.68 & \textbf{92.06} / \textbf{98.10} / \textbf{94.18} / \textbf{94.78} & \textbf{$+1.64^\ddagger$} & \textbf{$+1.10^\ddagger$} \\
 & \textbf{2-shot} & 92.00 / 97.69 / 93.57 / 94.42 & \textbf{93.65} / \textbf{98.30} / \textbf{94.63} / \textbf{95.53} & \textbf{$+1.65^\ddagger$} & \textbf{$+1.11^\ddagger$} \\
 & \textbf{4-shot} & 93.06 / 97.76 / 93.80 / 94.87 & \textbf{94.63} / \textbf{98.36} / \textbf{94.82} / \textbf{95.94} & \textbf{$+1.57^\ddagger$} & \textbf{$+1.07^\ddagger$} \\
\midrule
\multicolumn{6}{l}{\textit{(b) MVTec-LOCO~\cite{bergmann2022beyond} (5 Categories, 5 Seeds; Split Into Structural vs.\ Logical Anomalies Against Identical Normal Images; I-AUROC / Comp.)}} \\
MVTec-LOCO (Pooled All) & 1 / 4-shot & 62.52/66.57 \;\big|\; 66.46/68.02 & 59.67/\textbf{67.32} \;\big|\; \textbf{68.49}/\textbf{70.96} & $-2.85^\ddagger$ / $+2.04^\dagger$ & $+0.75^\dagger$ / $+2.94^\ddagger$ \\
MVTec-LOCO (\texttt{Structural}) & 1 / 4-shot & \textbf{75.09}/84.44 \;\big|\; 80.43/86.81 & 73.02/84.30 \;\big|\; \textbf{81.96}/\textbf{88.76} & $-2.08^\ddagger$ / $+1.52$ & $-0.13$ / $+1.95^\ddagger$ \\
MVTec-LOCO (\texttt{Logical}) & 1 / 4-shot & \textbf{53.17}/60.02 \;\big|\; 56.01/60.99 & 49.53/\textbf{60.88} \;\big|\; \textbf{58.33}/\textbf{64.30} & $-3.63^\ddagger$ / $+2.32^\dagger$ & $+0.86^\dagger$ / $+3.31^\ddagger$ \\
\midrule
\multicolumn{6}{l}{\textit{(c) Unseen Test-Time Corruptions (6 Categories: \texttt{cable, hazelnut, screw, pcb1, pcb2, pcb4}; Clean 1-Shot Ref., I-AUROC / Comp.)}} \\
Clean / Mean of 7 Corruptions & 1-shot & 90.20 / 93.05 \;\big|\; 90.43 / 93.03 & \textbf{93.45 / 95.23} \;\big|\; \textbf{93.10 / 95.02} & $+3.25$ / \textbf{$+2.67$} & $+2.18$ / \textbf{$+1.99$} \\
Gamma ($0.8/1.25$) \& Bright ($\pm 20\%$) & 1-shot & 90.20 / 93.03 \;\big|\; 90.36 / 93.07 & \textbf{93.43 / 95.22} \;\big|\; \textbf{93.26 / 95.19} & $+3.23$ / $+2.90$ & $+2.18$ / $+2.11$ \\
Gauss.\ Noise ($0.03/0.06$) \& Rot.\ ($\pm 15^\circ$) & 1-shot & 90.48 / 92.92 \;\big|\; 90.94 / 93.19 & \textbf{92.35 / 94.62} \;\big|\; \textbf{93.57 / 95.10} & $+1.88$ / $+2.63$ & $+1.70$ / $+1.92$ \\
\bottomrule
\end{tabular}\end{fitdatatable}
\end{table*}

\subsubsection{Why Unnormalized Local Euclidean Geometry?}
Table~\ref{tab:geometry_study} and Fig.~\ref{fig:sensitivity}(c) compare unit-sphere ($\Sp^{d-1}$), product ($\R_+ \times \Sp^{d-1}$, warped $\R_+ \times_r \Sp^{d-1}$), and local Euclidean ($\R^{D-s}$) manifolds on \texttt{seed=0}. Under the \emph{strictly matched} configuration $(s=48, M=8)$, replacing \texttt{Sphere-Local} with \texttt{Euclid-Local} immediately lifts All-27 I-AUROC by $+1.08\%$ ($94.85\% \to 95.93\%$), PRO by $+0.87\%$, and Composite by $+0.71\%$ ($95.74\% \to 96.45\%$), recovering \textbf{$+9.45\%$ I-AUROC on \texttt{screw}} ($75.43\% \to 84.87\%$) and \textbf{$+6.72\%$ on \texttt{pcb3}} ($83.81\% \to 90.53\%$) at half the scoring latency ($3.23$ vs.\ $6.74$\,ms). Combining local Euclidean geometry with tuned $(s=32, M=12)$ further lifts \texttt{DPLM-AD} to $96.50\%$ Composite ($87.15\%$ on \texttt{screw}, $91.10\%$ on \texttt{pcb3}). Unlike unit-sphere projection—which discards $2.4\text{--}2.7\times$ squared-norm contrast between foreground texture and low-norm background—or warped products requiring radial-angular tuning, unnormalized $\R^{D-s}$ preserves radial defect energy natively.

\subsubsection{Backbone Generalization, Stability, and FP16 Compression}
Table~\ref{tab:backbones_5seed} confirms gains across every tested backbone: \texttt{DINOv2-L/14} ($+0.74\%$ I-AUROC, $p=2.6\times 10^{-4}$), \texttt{DINOv2-G/14} ($+0.68\%$, $p=0.031$, cutting memory by $29.0\%$), and \texttt{DINOv3-L/16} ($+1.35\%$). On the shared \texttt{DINOv3-L/16} backbone, DPLM-AD incurs only $+0.69$\,ms end-to-end latency overhead ($41.10$ vs.\ $40.41$\,ms) and $6.91$\,MB FP32 ($3.43$\,MB FP16) state vs.\ $1.67$\,MB for SubspaceAD; crucially, unlocking this 304M backbone surpasses 1.1B \texttt{DINOv2-G/14} SubspaceAD ($+1.15\%$ I-AUROC, $p=0.012$; $+0.81\%$ PRO, $p=8.2\times 10^{-5}$) at \textbf{$3.93\times$ higher system speed} ($41.10$ vs.\ $161.61$\,ms) and $59.4\%$ lower FP32 state. Fig.~\ref{fig:sensitivity}(a)--(b) and Table~\ref{tab:diff_fp16} confirm broad hyperparameter stability ($0.11\%$ variation for $s\in[16,48]$, $0.07\%$ for $T\in[1,7]$, $0.28\%$ over $(w_g,\tau_g)\in\{3,5,7\}\times\{0.05,0.1,0.2\}$) and lossless FP16 quantization ($\le 0.0002\%$ max 5-seed macro $|\Delta|$).

\subsection{Extended Benchmarks, Split LOCO, and Qualitative Maps}
\label{sec:exp_robust_loco}

\begin{figure*}[t]
\centering
\includegraphics[width=0.68\textwidth]{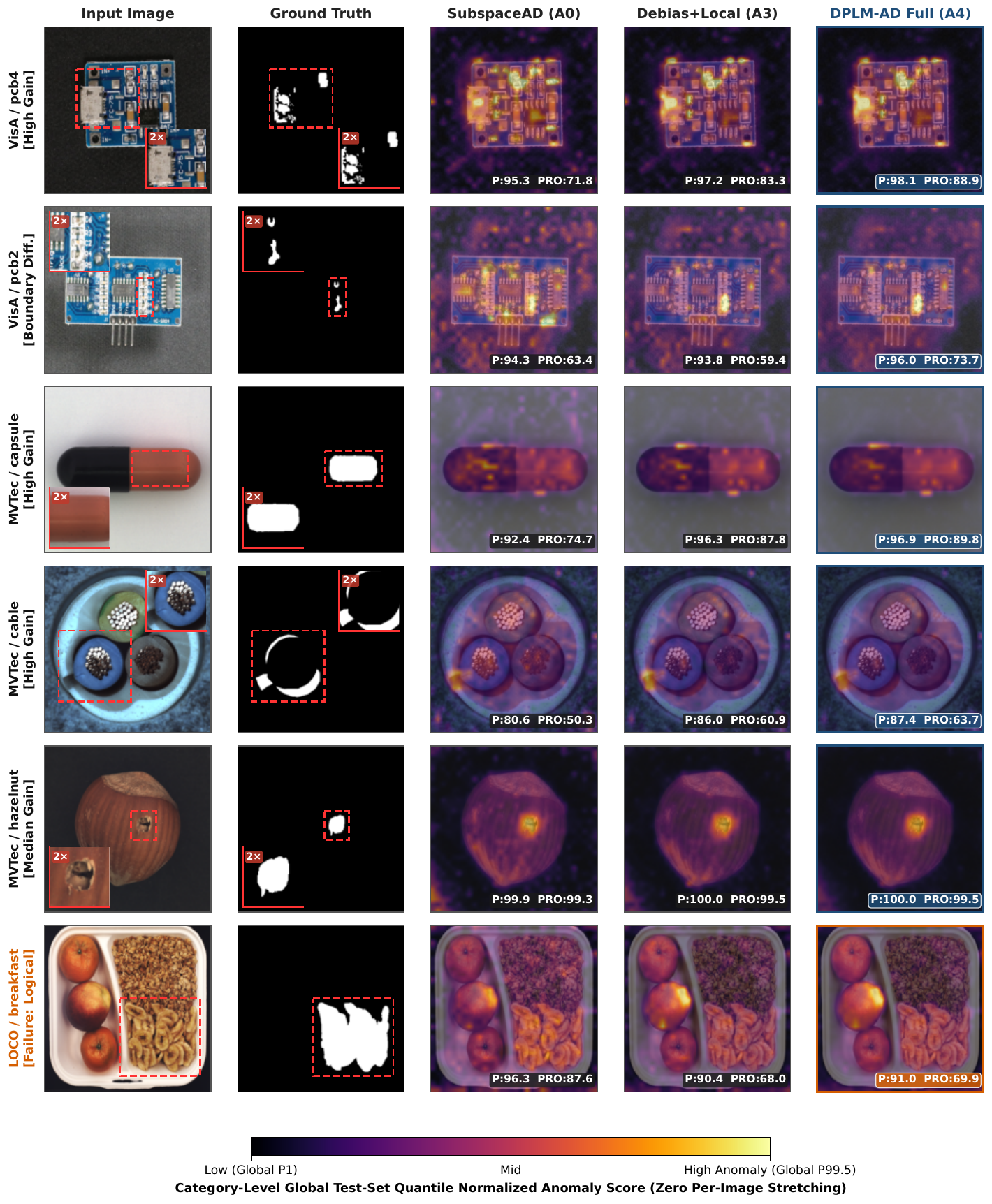}
\caption{Qualitative 1-shot localization (\texttt{seed=0}) under category-level global quantile calibration $[P_1, P_{99.5}]$ with transparent selection criteria: Rows~1, 3, and 4 show high-gain structural cases selected by combined P-AUROC/PRO improvement over A0; Row~2 (\texttt{pcb2/018.JPG}) illustrates the dual-space boundary case where Raw-Graph (A4, PRO $73.7\%$) outperforms both Gaussian (A3, $59.4\%$) and Deb-Graph ($70.3\%$); Row~5 (\texttt{hazelnut/009.png}) shows the anomalous image closest to the category-median PRO gain; Row~6 (\texttt{breakfast\_box/007.png}) shows a representative 1-shot logical failure where removing positional cues attenuates compartment-layout anomalies detected by A0.}
\label{fig:qualitative}
\end{figure*}

\subsubsection{MPDD, BTAD, All-36, Split LOCO, and Perturbations}
Table~\ref{tab:extra_robust} reports generalization without hyperparameter retuning:
\begin{itemize}
    \item \textbf{MPDD, BTAD, and All-36:} On multi-pose \textbf{MPDD}, DPLM-AD delivers $+3.78\% / +5.03\% / +4.37\%$ I-AUROC and $+1.91\% / +2.31\% / +2.00\%$ Composite across 1/2/4-shot. On \textbf{BTAD}, P-AUROC rises ($+0.08\%$ to $+0.12\%$) while PRO drops on unimodal \texttt{02} ($D_{\text{cent}}=0.247$). Across \textbf{all 36 structural categories}, DPLM-AD gains \textbf{$+1.64\% / +1.65\% / +1.57\%$ I-AUROC} and \textbf{$+1.10\% / +1.11\% / +1.07\%$ Composite} ($t$-test $p < 1.5\times 10^{-3}$, Comp.\ $p < 2\times 10^{-4}$; 1-shot Wilcoxon $p = 2.2\times 10^{-7}$, $33/36$ Comp.\ wins; $36/36$ P-AUROC wins at 4-shot).
    \item \textbf{MVTec-LOCO Structural vs.\ Logical Split (Table~\ref{tab:extra_robust}(b)):} Separating \texttt{Structural} and \texttt{Logical} anomalies against identical normal images clarifies our negative 1-shot result (Supplementary Table~S4). At 1-shot, removing positional bias degrades \texttt{Logical} I-AUROC more sharply than \texttt{Structural} ($-3.63\%$ vs.\ $-2.08\%$, led by fixed-layout \texttt{breakfast\_box}: $-15.94\%$ logical, $-8.71\%$ structural), while still lifting PRO ($+2.66\%$ / $+1.00\%$). By 4-shot, DPLM-AD surpasses SubspaceAD on both \texttt{Structural} ($+1.52\%$ I-AUROC, $+2.29\%$ PRO, $+1.95\%$ Comp.) and \texttt{Logical} ($+2.32\%$ I-AUROC $p=0.0255$, $+3.31\%$ Comp.); however, per-category breakdown shows both patch-level methods remain near chance on pure counting/layout categories (\texttt{pushpins} $45.5\%$, \texttt{screw\_bag} $43.1\%$), and our 4-shot logical gain stems chiefly from appearance-manifested violations (\texttt{juice\_bottle} wrong liquid/label $+9.63\%$, \texttt{splicing\_connectors} $+4.88\%$).
    \item \textbf{Test-Time Corruptions (Table~\ref{tab:extra_robust}(c)):} Across seven unseen photometric, noise, and $\pm 15^\circ$ perturbations with clean 1-shot references, DPLM-AD retains $99.78\%$ of clean Composite and leads SubspaceAD by $+2.67\%$ I-AUROC and $+1.99\%$ Composite.
\end{itemize}

\subsubsection{Qualitative Localization (High-Gain, Median, and Failure Cases)}
Fig.~\ref{fig:qualitative} displays 1-shot maps calibrated strictly by global test quantiles $[P_1, P_{99.5}]$ across high-gain, boundary-diffusion, category-median, and failure regimes. On multi-part structures (Rows~1--4), DPLM-AD suppresses background trace glow and sharpens micro-defect silhouettes (e.g., lifting single-specimen PRO from $71.8\%$ to $88.9\%$ on \texttt{pcb4} and from $59.4\%$ Gaussian / $70.3\%$ Deb-Graph to $73.7\%$ Raw-Graph on \texttt{pcb2/018.JPG}); Row~5 confirms clean typical behavior at the category-median gain on \texttt{hazelnut}, while Row~6 transparently illustrates our 1-shot logical-anomaly failure mode on \texttt{breakfast\_box}.

\section{Conclusion}
\label{sec:conclusion}

This work investigates the limitations of subspace-based normality modeling with dense vision foundation model features in training-free few-shot anomaly detection. Our analysis shows that the effectiveness of local subspace modeling depends not only on the diversity of normal appearance patterns but also on how positional information is encoded in the feature space. Although positional information can interfere with appearance-based clustering across images, it remains useful for spatial refinement within individual images. Based on these observations, we develop DPLM-AD, which models normal appearance using position-debiased local affine subspaces and refines anomaly scores through raw-feature-guided graph diffusion. Experimental results across five industrial benchmarks support the effectiveness of this design in improving anomaly detection and localization with limited computational overhead. More broadly, our findings suggest that different stages of anomaly detection can benefit from different feature representations, rather than relying on a single feature space throughout the entire pipeline.

\textbf{Limitations and Future Work:}
The proposed method has two main limitations. First, local subspace modeling may provide limited benefits when normal features exhibit a relatively homogeneous distribution or when the available reference samples are insufficient for reliable local subspace estimation. Although centroid dispersion and the sample-to-dimension ratio provide useful indicators, determining the appropriate number and rank of local subspaces remains an open problem. Second, removing positional information can reduce sensitivity to logical anomalies involving global spatial arrangements, particularly when only one normal reference is available. Future work will investigate adaptive subspace selection based on normal reference statistics and incorporate complementary layout modeling using the positional component.

\bibliographystyle{IEEEtran}
\bibliography{references}
\end{document}